\documentclass[acmtog]{acmart}
\acmSubmissionID{1187}

\usepackage{booktabs} 

\usepackage[ruled]{algorithm2e} 

\SetAlFnt{\small}
\SetAlCapFnt{\small}
\SetAlCapNameFnt{\small}
\SetAlCapHSkip{0pt}

\acmJournal{TOG}

\usepackage{multirow}
\usepackage{makecell}
\usepackage{numprint}
\usepackage{xr}
\usepackage{xcolor}

\definecolor{mycolor}{RGB}{0,0,0}
\definecolor{mycolor2}{RGB}{0,0,0}

\begin{document}
\title{HFH-Font: Few-shot Chinese Font Synthesis with Higher Quality, Faster Speed, and Higher Resolution}

\author{Hua Li}
\orcid{0009-0007-5678-9302}
\affiliation{%
  \institution{Wangxuan Institute of Computer Technology, Peking University}
  \city{Beijing}
  \country{China}}
\email{lhhh@pku.edu.cn}

\author{Zhouhui Lian}
\orcid{0000-0002-2683-7170}
\affiliation{%
  \institution{Wangxuan Institute of Computer Technology, Peking University}
  \city{Beijing}
  \country{China}}
\additionalaffiliation{%
  \institution{State Key Laboratory of General Artificial Intelligence, Peking University, Beijing, China}
  \city{Beijing}
  \country{China}}
\authornote{Corresponding author}
\email{lianzhouhui@pku.edu.cn}


\setcopyright{acmlicensed}
\acmJournal{TOG}
\acmYear{2024} \acmVolume{43} \acmNumber{6} \acmArticle{176} \acmMonth{12}\acmDOI{10.1145/3687994}

\begin{abstract}

The challenge of automatically synthesizing high-quality vector fonts, particularly for writing systems (e.g., Chinese) consisting of huge amounts of complex glyphs, remains unsolved. Existing font synthesis techniques fall into two categories: 1) methods that directly generate vector glyphs, and 2) methods that initially synthesize glyph images and then vectorize them. However, the first category often fails to construct complete and correct shapes for complex glyphs, while the latter struggles to efficiently synthesize high-resolution (i.e., 1024 $\times$ 1024 or higher) glyph images while preserving local details. In this paper, we introduce HFH-Font, a few-shot font synthesis method capable of efficiently generating high-resolution glyph images that can be converted into high-quality vector glyphs. More specifically, our method employs a diffusion model-based generative framework with component-aware conditioning to learn different levels of style information adaptable to varying input reference sizes. We also design a distillation module based on Score Distillation Sampling for 1-step fast inference, and a style-guided super-resolution module to refine and upscale low-resolution synthesis results. Extensive experiments, including a user study with professional font designers, have been conducted to demonstrate that our method significantly outperforms existing font synthesis approaches. Experimental results show that our method produces high-fidelity, high-resolution raster images which can be vectorized into high-quality vector fonts. Using our method, for the first time, large-scale Chinese vector fonts of a quality comparable to those manually created by professional font designers can be automatically generated.

\end{abstract}


%
%




%
%


\begin{teaserfigure}
\centering
  \includegraphics[width=\textwidth]{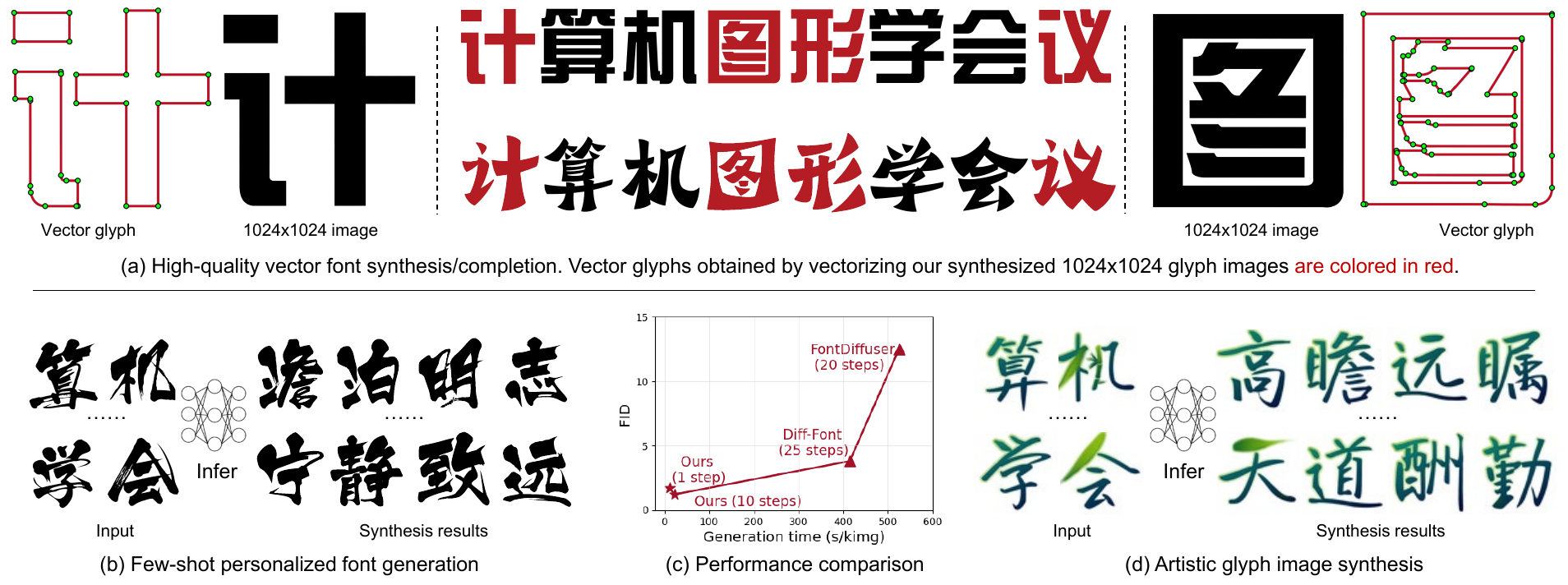}
  \caption{Our proposed HFH-Font aims to handle the task of few-shot Chinese font synthesis with higher quality, faster speed, and higher resolution. (a) Taking a subset of characters in the desired style as inputs, our method outputs high-fidelity, high-resolution glyph images of the remaining characters, which can be vectorized into high-quality vector fonts. Applications to (b) few-shot personalized font generation and (d) artistic glyph image synthesis can also be fulfilled with visually pleasing synthesis results. (c) The diffusion-based generation process is further accelerated to 1-step inference, surpassing current state-of-the-art diffusion model-based font synthesis methods in both generation quality and efficiency. Please zoom in for better inspection.}
  \label{fig:teaser}
\end{teaserfigure}

\maketitle

\section{Introduction}

Font synthesis is the task of automatically generating an entire font library using only limited glyph examples (see Fig.~\ref{fig:teaser}). Automatic font generation alleviates the burden of extensive, time-consuming manual design, and therefore has great application potential and has received increasing attention in recent years. Especially for glyph-rich writing systems such as Chinese, Japanese, or Korean, which contain tens of thousands of characters, a fast and effective few-shot font generation method not only speeds up the design of high-quality commercial font products, but also allows for applications in personalized handwriting generation, scene text image editing, data augmentation for optical character recognition (OCR), etc.

Existing font synthesis methods generally fall into two categories: directly targeting font generation in the vector modality~\cite{wang2021deepvecfont,wang2023deepvecfont,thamizharasan2023vecfusion}, or generating raster images before vectorizing them into vector fonts. This paper focuses on the investigation of the second category, i.e., we aim to develop a system that takes glyph images of a subset of characters in the desired style as inputs and produces glyph images of the rest of the characters as outputs; using the synthesized high-resolution glyph images, high-quality vector fonts can be obtained by employing well-studied image vectorization techniques. Compared to methods that directly target vector font synthesis, although dealing with high-resolution images involves high computational complexity and manual interventions are still needed in the vectorization process for existing approaches, the image modality that consists of structural information is much easier to process by many state-of-the-art generative models. On the other hand, font generation methods that directly process the vector modality are currently only capable of dealing with simple shapes, limiting their application in font generation for writing systems containing glyphs with high topological complexity, e.g., Chinese. As shown in Fig.~\ref{fig:teaser}(a), we can see that, once high-resolution (i.e., 1024 x 1024 or higher) glyph images with well-preserved local details can be synthesized, it is possible to generate high-quality vector glyphs which are indistinguishable from those manually produced by professional font designers.


Despite recent advances, few-shot font generation remains an unsolved and challenging task due to the complexity of character structures and the creative, diverse nature of font design. Most previous approaches are built upon Generative Adversarial Networks (GANs)~\cite{goodfellow2020generative} and adopt the style-content disentanglement scheme in EMD~\cite{zhang2018separating} to train an image-to-image translation network in an adversarial manner. However, GANs are known to be hard to train, difficult to scale, suffer from mode collapse, and lack diversity, limiting their potential in generation quality. Moreover, a deterministic generative network is unable to generate multiple results given specific inputs, while an ideal system gives users control over the generation process and freedom to choose from diverse generated results. On the other hand, 
diffusion models have been shown to produce better results compared to previously state-of-the-art GAN-based methods~\cite{dhariwal2021diffusion}, can be easily scaled to larger datasets, and are able to generate diverse results given specific conditions. However, the most prominent drawback of diffusion models is their iterative inference process that prevents the practical use of such models. Moreover, properly integrating prior knowledge of glyphs into diffusion models also requires delicate and specific designs.

In this paper, we propose HFH-Font, a diffusion model-based font synthesis method that supports high-quality one-step few-shot font generation as well as generation in high resolution (e.g., 1024 $\times$ 1024). 
\textcolor{mycolor}{For glyph-rich systems like Chinese, which contains tens of thousands of characters (eg. 6,763 for the GB2312 standard, 87,887 for the GB18030-2022 standard), extremely few-shot generation is often ill-posed since hundreds of input references are usually required to cover all possible components. On the other hand, it is also desired for the model to learn a global representation and make assumptions of the missing knowledge when limited references are available in real-world scenarios. 
Therefore, to fully exploit the information of sufficient amount of references (eg., hundreds), which most extremely few-shot learning methods do not support, motivated by FsFont~\cite{tang2022few}, we leverage an attention-based conditioning module to enable reuse of component-level style information, which can learn to effectively transfer diverse and complex font styles to target glyphs when combined with the strong generative capability of latent diffusion models. 
Additionally, to allow for smaller input reference sizes (eg., less than 10) in real-world scenarios, we introduce a reference selection strategy to equip the model with the ability to extract both global and local style information.} 
To tackle the problem of slow inference in the diffusion process, we incorporate the Score Distillation Sampling technique~\cite{poole2022dreamfusion} in the text-to-3D generation literature into our framework. In this manner, we achieve efficient one-step generation with little loss of quality. To further enable generation in higher resolution, we design a super-resolution process where the loss of detailed styles in the lower-resolution regime can be recovered during upscaling, and the gap between the image and vector domains can be mitigated.

Closely related to our work are Diff-Font~\cite{he2022diff} and FontDiffuser~\cite{yang2023fontdiffuser}, which apply pixel-space diffusion models~\cite{ho2020denoising,dhariwal2021diffusion} to the glyph image synthesizing task and utilize either stroke vectors or multi-scale content features to better preserve glyph structures. Our approach differs in the following aspects: 1) we use a latent-space diffusion model instead of a pixel-space diffusion model, markedly reducing both training and inference costs; 2) we conduct component-level style extraction and transformation, obtaining substantially superior style accuracy and consistency; 3) we utilize diffusion distillation and super-resolution techniques for high-resolution, one-step image generation. Our proposed method significantly improves both the generative quality and practical feasibility, yielding superior and impressive results.

To sum up, major contributions of this paper are as follows:

\begin{itemize}
\item We propose HFH-Font, a diffusion model-based font generation framework that supports high-quality font generation in as few as one step, as well as generation in high resolution (e.g., 1024 $\times$ 1024) which facilitates further vectorization into high-quality vector fonts. 

\item We introduce a component-aware conditioning module and a reference selection strategy that enable the model to deal with different levels of style information and adapt to different reference sizes, allowing for trade-off between style fidelity and input reference size.

\item We conduct extensive experiments, including a user study with professional font designers, to demonstrate the superiority of our method to the state of the art. According to both the quantitative evaluation and user studies, our method outperforms existing approaches by a considerable margin. In addition, we showcase our model’s ability to transfer both shape and texture styles through application to artistic glyph image synthesis.
\end{itemize}


\section{Related Work}

\subsection{Font Generation}

\subsubsection{Font Generation in Vector Modality}

Significant work has been done in generation of vector fonts. Early work such as~\cite{campbell2014learning} proposes to learn a font manifold in which new fonts can be synthesized. Recent methods employ deep generative models targeting sequential data for the task of vector font generation. SketchRNN~\cite{ha2017neural} applies bi-directional RNNs to generate stroke-based drawings. FontRNN~\cite{tang2019fontrnn} utilizes RNNs to synthesize the writing trajectories of Chinese characters. DeepSVG~\cite{carlier2020deepsvg} proposes a transformer-based VAE based on hierarchical representation of SVG for generation of vector icons. DeepVecFont~\cite{wang2021deepvecfont} and DeepVecFont-v2~\cite{wang2023deepvecfont} adopt techniques in image synthesis and sequence modeling to fully exploit dual modality information.
\textcolor{mycolor}{DualVector~\cite{liu2023dualvector} proposes a dual-part font representation for unsupervised shape modeling, followed by a contour refinement procedure for better details. VecFontSDF~\cite{xia2023vecfontsdf} utilizes the SDF representation to reconstruct and synthesize high-quality vector fonts.} 
VecFusion~\cite{thamizharasan2023vecfusion} uses an image as well as vector diffusion model in a cascaded manner for the generation of complex shapes and diverse styles. Due to the limited ability of dealing with long and complex sequences, these methods are only capable of handling simple shapes like English characters or simple Chinese characters with fewer strokes and shorter drawing paths. The generated styles also lack diversity. Although impressive results for English font synthesis have been shown in~\cite{thamizharasan2023vecfusion}, it can not be directly applied to handle Chinese characters that contain longer sequences and more topological complexity. To address this problem, EasyFont~\cite{lian2018easyfont} utilizes shallow neural networks to learn and recover a user’s overall
handwriting styles and detailed handwriting behaviors, and thus automatically generate large-scale Chinese handwriting fonts in the user's personal style; CVFont~\cite{lian2022cvfont} uses a layout prediction module to learn the layout of corresponding components of target glyphs. But they do not generalize to unseen fonts, still require considerable number of references, and cannot handle diverse styles.

\subsubsection{Font Generation in Image Modality}

As mentioned above, although glyph image is a much less efficient and lossy representation of fonts and requires additional vectorizing to obtain vector fonts, it possesses desirable qualities such as easy management, ability to handle complex glyphs and styles, and better generation quality thanks to rapid growth in the field of image generation. Font generation in the image modality has seen great advances in the deep learning era. Early approaches such as zi2zi~\cite{zi2zi} and DCFont~\cite{jiang2017dcfont} that treat the task as an image translation problem are built upon the pix2pix~\cite{isola2017image} framework and require per-font fine-tuning at test time. EMD~\cite{zhang2018separating} uses separate style and content encoders to achieve style-content disentanglement, enabling generalization to unseen fonts. Subsequent works employ this disentanglement scheme with refinements, e.g., ZiGAN~\cite{wen2021zigan} maps style features to the Hilbert space and aligns the feature distributions, DG-Font~\cite{xie2021dg} introduces deformation convolution to the generator, CF-Font~\cite{wang2023cf} proposes a content fusion module to narrow the gap between the reference and target fonts, etc. 

For the generation of highly structured CJK characters specifically, notable work utilizes prior domain knowledge such as stroke order~\cite{zeng2021strokegan} or stroke/component decomposition~\cite{jiang2019scfont, tang2022few} to guide the generation process at stroke-level or component-level. Stroke-GAN~\cite{zeng2021strokegan} conditions the discriminator on stroke count information to reduce stroke-level errors.
\textcolor{mycolor}{DM-Font~\cite{cha2020few} is the first to utilize the compositionality of a script by encoding component-wise styles into the dynamic memory.
MX-Font~\cite{park2021multiple} employs a multi-head design and each head is specialized for different local concepts.}
CG-GAN~\cite{kong2022look} trains a component predictor to better supervise the generator during adversarial training. 
FsFont~\cite{tang2022few} constructs a character-reference mapping and utilizes cross-attention to guide the network to focus on relevant components in the reference images with each source character. 
\textcolor{mycolor}{VQ-Font~\cite{pan2023few} uses a quantization-based variational autoencoder to automatically extract components and an attention-based module to transfer local styles.} 


\subsection{Diffusion Models}

Diffusion-based generative models~\cite{sohl2015deep,ho2020denoising} are a family of generative models that have recently achieved unparalleled performance in many image synthesis tasks \cite{podell2023sdxl} as well as tasks targeting other modalities such as video~\cite{blattmann2023align}, audio~\cite{kong2020diffwave}, text~\cite{li2022diffusion}, etc. Latent diffusion models (LDMs)~\cite{rombach2022high} particularly have been widely used in a variety of conditional generation tasks including image editing~\cite{brooks2023instructpix2pix}, text-to-image generation~\cite{podell2023sdxl}, etc, the reduced computational cost of which makes it possible for diffusion models to be applied in real-life applications. Cascaded diffusion models~\cite{ho2022cascaded} are proposed to successively upsample generated images to higher resolutions through consecutive super-resolution models and has been successfully used in multiple tasks targeting high-resolution image generation~\cite{saharia2022photorealistic}.


Generation of text within images has been a major challenge for generative models. Efforts have been made in scene text generation and editing~\cite{yang2023glyphcontrol,ma2023glyphdraw,tuo2023anytext} and handwriting generation~\cite{luhman2020diffusion,gui2023zero} that attempt to utilize the strong generative capability of diffusion models for text generation within images. We focus on gray-scale glyph image generation where more complex font styles and finer character details are dealt with, laying the foundation for further applications. Recently, Diff-Font~\cite{he2022diff}, FontDiffuser~\cite{yang2023fontdiffuser}, and QT-Font~\cite{liu2024qtfont} have attempted to apply diffusion models to this task and have achieved impressive results. Although they are able to generate sharp results thanks to the generative ability of diffusion models, the pre-trained style features or style contrastive refinement strategy they use often fail when facing unseen, complex font styles and the pixel-space multi-step diffusion they employ limits their methods’ practical use and scalability to higher resolution. In contrast, we provide support for one-step as well as high-resolution generation and incorporate prior knowledge on component decomposition into our system which significantly benefits the learning and transferring of font style. 

\subsection{Speeding Up Diffusion Models}

The multi-step iterative inference process is a primal issue with diffusion models and accelerating the inference process has become a main focus in the field. Accelerating can be done by either reducing the number of sampling steps or improving network architecture. We focus on the former and leave the optimization of network architecture for the task of font generation for future work. 

Advanced samplers such as DDIM~\cite{song2020denoising}, DPM-Solver~\cite{lu2022dpm}, and DEIS~\cite{zhang2022fast} have emerged that are able to reduce the number of sampling steps to around 10-20. Although they do not require additional training, they fail drastically in the few-step regime, i.e., single or double step. To accomplish few-step diffusion sampling, distillation methods are proposed to distill the knowledge of multi-step teacher models into few-step student models with the minimum quality degradation. ~\cite{luhman2021knowledge} proposes to straightforwardly distill diffusion teachers into one-step student models that learn to output the deterministic DDIM results sampled by the teacher models. To circumvent the problem of having to generate large amounts of synthetic training data offline, progressive distillation~\cite{salimans2022progressive,meng2023distillation} is proposed, where the number of sampling steps is halved at each iteration by training the student model to match the 2-step output of the teacher model. Alternatively, ~\cite{song2023consistency} introduces consistency models that achieves single-step generation by enforcing self-consistency between two adjacent points along the same deterministic diffusion trajectory and has later been applied to speed up latent diffusion models~\cite{luo2023latent}. 

Recently, several works turn to borrow from the text-to-3D literature and have achieved state-of-the-art results in single-step diffusion generation~\cite{sauer2023adversarial,yin2023one}. The techniques of utilizing knowledge from pre-trained large-scale diffusion models to guide 3D models in producing realistic results, e.g., Score Distillation Sampling~\cite{poole2022dreamfusion,sauer2023adversarial} and Variational Score Distillation~\cite{wang2023prolificdreamer,yin2023one,hoang2023swiftbrush}, have been found effective for guiding the learning of a single-step generation network with the aid of an additional regression or adversarial loss. In this paper, we find that a simple score distillation loss works sufficiently and effectively for our font synthesis task.


\begin{figure*}[!ht]
\centering
\includegraphics[width=0.8\paperwidth]{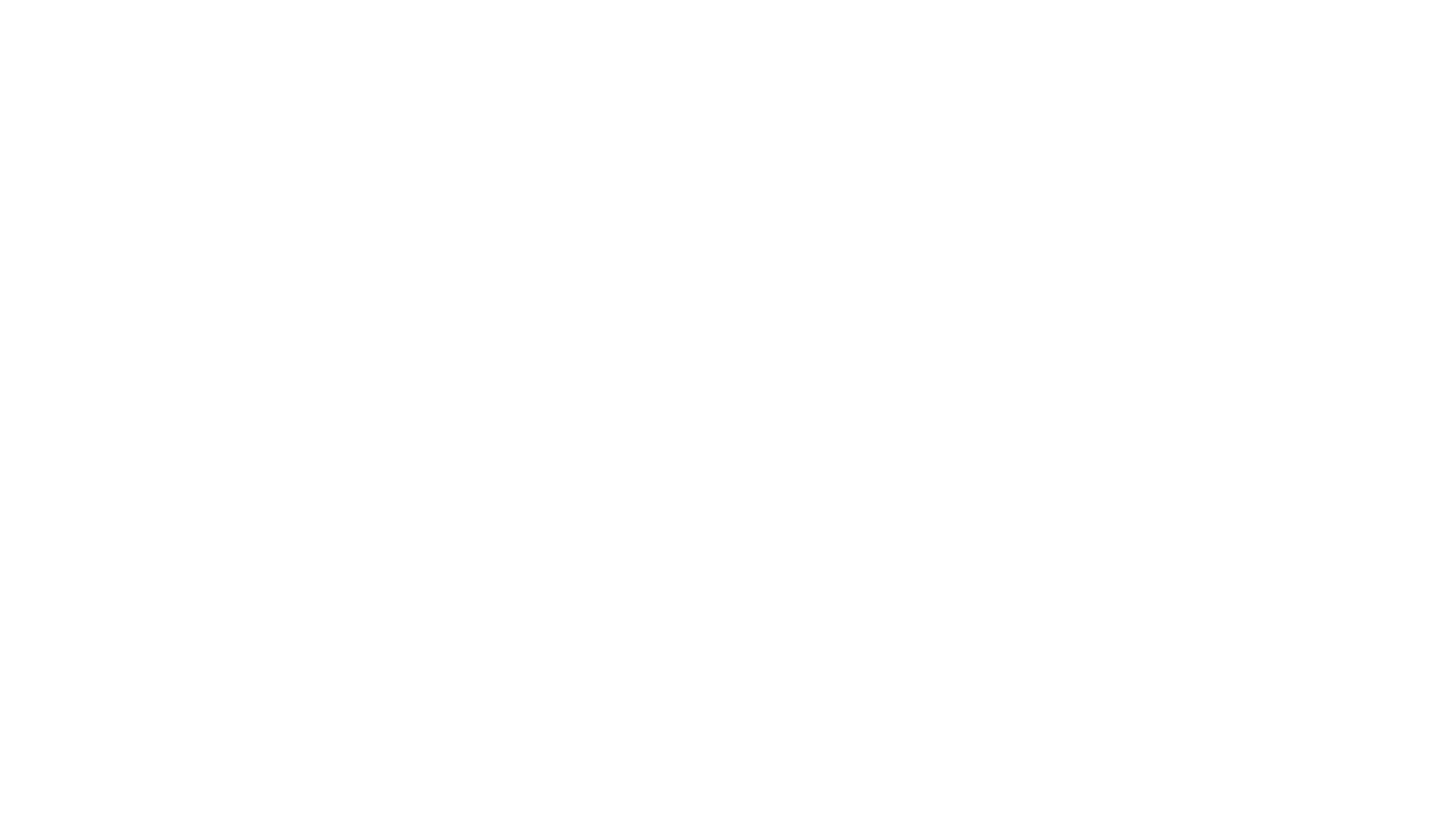}
\caption{An overview of our method. The three segments denote three parts of our framework. Weights from the Stage A low-resolution model are used to initialize the training of Stage B1 and Stage B2, which are parallel to each other. Note that the colored images are only for visualization; only gray-scale glyph images are input into the network.}
\label{fig:overview}
\end{figure*}

\section{Method}

The overall framework of our method is shown in Fig.~\ref{fig:overview}, which consists of three parts: a conditional latent diffusion model with component-aware encoders that is used to generate high-fidelity low-resolution results (i.e., 64 $\times$ 64), a diffusion distillation branch for distilling the trained model into a one-step generation model, and a super-resolution model that is used to generate high-resolution glyphs in a cascaded manner. The three parts will be illustrated in detail in Section~\ref{sec:ldm}, Section~\ref{sec:distill}, and Section~\ref{sec:high}, respectively.

\subsection{Conditional LDM with Component-aware Conditioning}
\label{sec:ldm}

\subsubsection{Latent diffusion model for font generation}

The main idea of diffusion models is to gradually perturb real data with increasing levels of tractable (e.g., Gaussian) noise until the real signal is transformed into the pure noise, and learn a denoising function to gradually denoise the pure noise back to the real data. Specifically, let $x_0\sim q(x_0)$ be a sample from the real data distribution $q(x)$, at each noise level $t\in\{1,…,T\}$, $x_0$ is perturbed with the Gaussian noise $\epsilon$ using the noise schedule $\{\alpha_t\}, \{\sigma_t\}$:

\begin{equation}
x_t=\alpha_tx_0+\sigma_t\epsilon,\ \  \epsilon\in\mathcal{N}(0,1),\ \ \alpha_t^2+\sigma_t^2=1,\ \
\alpha_t,\sigma_t\ge0.
\end{equation}

We follow the practice in DDPM~\cite{ho2020denoising} and train a $\epsilon$-prediction network $\epsilon_\theta(x_t, t)$ to predict the noise added to $x_0$ to obtain $x_t$ using the variational bound, which can be simplified to the loss $\mathcal{L}=\mathbb{E}_{x_0,\epsilon,t}[\parallel\epsilon-\epsilon_\theta(x_t,t)\parallel_2^2]$, where the weighting function is simply discarded. We base our method on latent diffusion models~\cite{rombach2022high}, a variant of the original DDPM where the perturbing and denoising processes are done in the latent space of an autoencoder. Let $\mathcal{E}$ be the encoder in LDM that compresses the original data into lower-dimensional representations and $\mathcal{D}$ be the decoder, by substituting $x_0$ with $z_0=\mathcal{E}(x_0)$, we obtain the diffusion loss in the latent space:

\begin{equation}
\mathcal{L}=\mathbb{E}_{x_0,\epsilon,t}[\parallel\epsilon-\epsilon_\theta(z_t,t)\parallel_2^2], \quad z_t=\alpha_t z_0+\sigma_t\epsilon.
\end{equation}

Given a content reference image $x_c$ rendered in the source font style (e.g., Heiti) and a set of $k$ style reference images $X_s=\{x_s^1,…,x_s^k\}$ in the desired font, our goal is to generate a glyph image in the desired font that aligns with the content in $x_c$. It is essentially an image-to-image translation task and a straightforward way to achieve this is to train a conditional diffusion model conditioned on the input glyph images. Our method adopts a component-aware module $\mathcal{C}$, which we will later describe in depth, to encode the information of input content and style reference glyph images into the conditioning vector $y=\mathcal{C}(x_c,X_s)$. Thereby, the new loss function is defined as:

\begin{equation}
\mathcal{L}=\mathbb{E}_{z_0,\epsilon,t}[\parallel\epsilon-\epsilon_\theta(z_t,t,\mathcal{C}(x_c,X_s))\parallel_2^2].
\end{equation}

In addition, in the multi-step diffusion sampling scenario, classifier-free guidance is frequently used to trade diversity for quality by leading the samples towards higher-density regions given certain input conditions. We follow the practice in~\cite{brooks2023instructpix2pix} of conducting classifier-free guidance for multiple conditioning. The guided score estimate is computed by:

\begin{equation}
\begin{split}
\tilde{\epsilon}_\theta(z_t,t,\mathcal{C}(x_c,X_s))&=\epsilon_\theta(z_t,t,\mathcal{C}(\varnothing,\varnothing))
\\
&+s_c\cdot(\epsilon_\theta(z_t,t,\mathcal{C}(x_c,\varnothing))-\epsilon_\theta(z_t,t,\mathcal{C}(\varnothing,\varnothing)))
\\
&+s_s\cdot(\epsilon_\theta(z_t,t,\mathcal{C}(x_c,X_s))-\epsilon_\theta(z_t,t,\mathcal{C}(x_c,\varnothing))),
\end{split}
\end{equation}
where the two guidance scales $s_c$ and $s_s$ control how strongly the generated samples correspond to content and style conditions, respectively. Null condition $\varnothing$ is done by setting input images to all-zero vectors. Following~\cite{brooks2023instructpix2pix}, we randomly set only $x_c=\varnothing$ for 5\% of the input conditions, only $X_s=\varnothing$ for 5\% of the input conditions, and both for 5\% of the input conditions.

\subsubsection{Component-aware conditioning}
\label{sec:ldm_comp}

As we know, the style of a font exists at different structural levels of a glyph image and different characters share different common style information. To be specific, global styles (e.g., size, aspect ratio, stroke thickness, overall spatial layout, etc) are ubiquitous among all characters, stroke-level styles (e.g., brush details near the end of certain strokes, sloping of certain strokes, etc) are shared among characters containing common strokes, and component-level styles (e.g., special designs of certain radicals) are only shared among characters with mutual components.

\begin{figure}[!t]
\centering
\includegraphics[width=0.4\paperwidth]{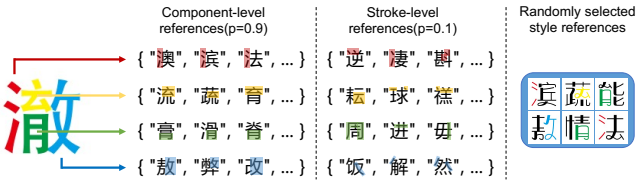}
\caption{Our reference selecting procedure that enables the model to deal with different levels of style information.}
\label{fig:ref_selection}
\end{figure}

Therefore, ideally, we would like our model to be able to focus on different relevant parts of the style references at different structural levels of the input content. Motivated by FsFont~\cite{tang2022few}, we resort to multi-head cross-attention to achieve such intention. Specifically, our component-aware conditioning module consists of a style encoder $E_s$, a content encoder $E_c$, and a cross-attention module. The style encoder separately encodes all $k$ style reference images $x_s^i,i=1,…,k$ into style features $f_s^i\in\mathbb{R}^{d_s\times h_s\times w_s},i=1…,k$, and the content encoder encodes the content reference image $x_c$ into the content feature $f_c\in\mathbb{R}^{d_c\times h_c\times w_c}$. We represent each of the style feature as $h_s\times w_s$ tokens of dimension $d_s$ and the content feature as $h_c\times w_c$ tokens of dimension $d_c$, and performs cross-attention:
\begin{equation}
\begin{aligned}
y=Attention(Q,K,V)=softmax(\frac{QK^T}{\sqrt{d}})\cdot V,\\
Q=W_Q^{(i)}\cdot \tilde{f}_c,\ K=W_K^{(i)}\cdot \tilde{f}_s,\ V=W_V^{(i)}\cdot \tilde{f}_s,
\end{aligned}
\end{equation}
where $W_Q^{(i)}\in\mathbb{R}^{d\times d_c}$, $W_K^{(i)},W_V^{(i)}\in\mathbb{R}^{d\times d_s}$ are projection matrices for the $i$-th head, $\tilde{f}_c$ is the flattened version of $f_c$, and $\tilde{f}_s$ is the flattened version of the aggregated style feature $f_s=concatenate(f_s^1,...,f_s^k)$. In this way, through attention mechanism, each fine-grained position on the content feature maps learns to attend to relevant positions on the style feature maps, and the relevant style features are gathered to form the final conditioning feature of size $d\times h_c\times h_c$. 

To guide such learning process, we explicitly feed our style encoder with carefully chosen style references that correspond to the components and strokes the content character contains.~\cite{lian2018easyfont} defines $m=1,032$ categories of components and $n=339$ categories of fine-grained strokes for Chinese characters, and constructs several character sets of different sizes that cover different ratios of components and strokes. We use the OptSet in~\cite{lian2018easyfont} that consists of 775 commonly used Chinese characters as our full style reference set since it covers all components that appear in Chinese characters from the GB2312 official standard. For each character $c$ in the training set, suppose $c$ consists of $m_c$ components that belong to the component category $comp_{c,i}\in\{1,…,m\},i=1…,m_c$, and the $i$-th component consists of $n_{c,i}$ strokes that belong to the stroke category $stroke_{c,i,j}\in\{1,…,n\},j=1,…,n_{c,i}$. 

Next, we try to establish the selection of style references. The number of style references $k$ is set to 6 since Chinese characters are typically decomposed into at most 6 components. We build a component-level character-reference mapping and a stroke-level character-reference mapping that assign a sub-reference set $X_{comp_{c,i}}$ to each component $comp_{c,i}$ in a character $c$. For the former mapping, $X_{comp_{c,i}}$ consists of all characters from the reference set that contains the component $comp_{c,i}$; for the latter, $X_{comp_{c,i}}$ consists of all characters from the reference set that contains any stroke from the stroke set $\{stroke_{c,i,j},j=1,…,n_{c,i}\}$. During training, we randomly select one character from each $X_{comp_{c,i}}$ to form the style reference set of character $c$ and then randomly fill the rest of the $k$ positions. The two mappings are used with a probability of $1-p$ and $p$, respectively. We use OptSet as the style reference set during training; during testing, new mappings are constructed according to the given reference set. See Fig.~\ref{fig:ref_selection} for an illustration of the reference selection procedure.

Although our style encoder has only seen OptSet during training, it is able to generalize to reference glyph images outside of OptSet, meeting the requirements of real-world application scenarios where arbitrary style references are available. On the other side, although we leave OptSet out of training images for simplicity, our model is still capable of generating characters in OptSet. Please refer to our supplemental materials for synthesized glyph images of Chinese characters in OptSet using style references outside of OptSet.

\subsection{One-step Generation via Score Distillation Sampling}
\label{sec:distill}

An image diffusion model usually requires tens of sampling steps in order to produce valid, high-quality outputs. We empirically find that for our model, sampling 10 steps using a simple DDPM sampler (i.e., 1-order SDE sampler) gives sufficiently good results. This is likely due to the reduced complexity of our target glyph image domain compared to the regular image domain; nevertheless, 10-step inference is still far from fast, real-time generation. In this section, we propose to distill the above trained model into a one-step generation model using a strategy based on Score Distillation Sampling~\cite{poole2022dreamfusion}.

Score Distillation Sampling (SDS) is originally proposed for the task of text-to-3D generation, where the knowledge of a pre-trained 2D text-to-image diffusion model is utilized to guide a 3D model to render plausible 2D images from random angles, in turn achieving effective learning of the 3D model. We find that this distillation process can also be employed to reduce the sampling steps of our model, without the need for an additional auxiliary regression~\cite{yin2023one} or the adversarial loss~\cite{sauer2023adversarial}. 

The distillation procedure is shown as Stage B1 in Fig.~\ref{fig:overview}. It involves a frozen trained teacher model $\epsilon_\theta(z_t,t,y)$ and a one-step generation student model with weights initialized from the teacher model as $\epsilon_\phi(z_T,y)=\epsilon_\theta(z_T,T,y)$. The conditioned time step in the student model is fixed to $T$ to ensure that the model starts from pure noise during inference~\cite{sauer2023adversarial,lin2024common}. During training, the student model produces samples $\hat{z}_\phi$ from the input $z_T=\alpha_T z_0+\sigma_T\epsilon$ and the condition $y$. Similar to how the original SDS method treats 2D images rendered from a 3D model, we aim to optimize the parameter $\phi$ so that $\hat{z}_\phi$ looks like a plausible sample from the teacher model. This is achieved by moving the sample towards the high-density region, which is implicitly defined by the teacher model at each time step $t$. The distillation is equivalent to optimizing the following objective:
\begin{equation}
\mathcal{L}_{SDS}=\mathbb{E}_{t,y}[\frac{\sigma_t}{\alpha_t}w(t)D_{KL}(q(z_t|\hat{z}_\phi;y,t)\parallel p_\theta(z_t|y,t))],
\end{equation}
where $q(z_t|\hat{z}_\phi;y,t)$ is the marginal distribution defined in the forward process (i.e., $z_t=\alpha_t \hat{z}_\phi+\sigma_t\epsilon$) and $p_\theta(z_t|y,t)$ is the marginal distribution at time step $t$ implicitly described by the teacher model.

Since estimating noise added to a data sample is equivalent to estimating the score of the perturbed data distribution up to some constant factors~\cite{song2020score}, the gradient of the above object can be approximated by:

\begin{equation}
\begin{split}
\nabla_{\phi}\mathcal{L}_{SDS}&=\mathbb{E}_{t,y}[\frac{\sigma_t}{\alpha_t}w(t)\nabla_\phi D_{KL}(q(z_t|\hat{z}_\phi;y,t)\parallel p_\theta(z_t|y,t))]
\\
&=\mathbb{E}_{t,\epsilon,y}[w(t)(\epsilon_\theta(z_t,t,y)-\epsilon)\frac{\partial \hat{z}_\phi}{\partial \phi}],
\end{split}
\end{equation}
where the noise estimation can be substituted by the classifier-free guided version. We only update diffusion weights, i.e., the conditioning weights are fixed during training. $w(t)$ is set to be uniform across all $t$ in our experiment. \textcolor{mycolor}{Note that unlike the method proposed in \cite{sauer2023adversarial} where the distillation loss is computed in the pixel space, our method operates in the latent space.}  

\subsection{Towards Higher Resolution}
\label{sec:high}

The motivation of moving the generation process towards higher resolution is two-fold: 1) a lot of information, such as style details or densely placed strokes, is lost at low resolution (e.g., 64 $\times$ 64, see Fig.~\ref{fig:highres_detail}), therefore an upsampling procedure is needed to recover such lost information; 2) the gap between raster images and vector fonts can be narrowed through upsampling, i.e., if we can generate sharp, correct, and high-quality images at the resolution 1024 $\times$ 1024 or higher, vectorizing can be done with negligible losses (see Fig.~\ref{fig:vector_glyphs} for some vectorization results of our generated glyph images compared to the corresponding ground-truth vector glyphs).



We achieve the generation of high-resolution glyph images by consecutively upsampling low-resolution results to higher resolutions using a cascade of conditional super-resolution diffusion models, i.e., from 64 $\times$ 64 to 256 $\times$ 256, and then from 256 $\times$ 256 to 1024 $\times$ 1024. Cascaded diffusion models~\cite{ho2022cascaded} have been used in the generation of high-resolution images with great success~\cite{saharia2022photorealistic}. The noise conditioning augmentation procedure is crucial for correcting artifacts generated by lower-resolution models. Instead of directly applying other existing image super-resolution models, we use the same framework adopted in Stage A, i.e., a latent diffusion model coupled with the component-aware conditioning module, for our super-resolution modules with two modifications: 1) we insert the given low-resolution image into the model by replacing the content reference image with the low-resolution image as well as concatenating it with input noisy latents $z_t$; in this way, similar to what cross-attention in the conditioning module is intended to do for the low-resolution model, we guide the attention at each position of the low-resolution image to relevant positions in style references in order to recover the lost details; 2) to perform noise conditioning augmentation, we encode the low-resolution images to the latent space, corrupt them with Gaussian noise, and then decode them back to the image space to serve as noise-augmented low-resolution image inputs; in addition, we add an extra condition $s$, the level of noise added during noise conditioning augmentation, to the diffusion U-Net model; 3) to deal with the increase of memory usage that comes with higher resolution, we add extra downsampling layers to our style and content encoders; the detailed network configurations are provided in the supplemental materials.

It should be pointed out that we use the weights from the low-resolution generation model to initialize our super-resolution models. This is because the style and content encoders there already possess knowledge of extracting and transferring desired styles.

\begin{table*}[!t]
\caption{Quantitative evaluation on unseen fonts.}
\label{tab:metrics}
\begin{minipage}{\textwidth}
\begin{center}
\begin{tabular}{c|c|cccccc|cccccc}
  \toprule

\multirow{2}{*}{Method} & \multirow{2}{*}{\makecell{Train\\Set}} & \multicolumn{6}{c|}{Unseen Fonts Seen Characters} & \multicolumn{6}{c}{Unseen Fonts Unseen Characters} \\ &\  &\  RMSE$\downarrow$ & SSIM$\uparrow$ & LPIPS$\downarrow$  & FID$\downarrow$ &  Acc(C)$\uparrow$ & Acc(S)$\uparrow$ & RMSE$\downarrow$ & SSIM$\uparrow$ & LPIPS$\downarrow$  & FID$\downarrow$ & Acc(C)$\uparrow$ & Acc(S)$\uparrow$ \\ \midrule

  \multirow{2}{*}{\makecell{MX-Font}}  
  &Small   & 0.364  
           & 0.387  
           & 0.251 
           & 21.955  
           & 0.984 
           & 0.082 
           & 0.365
           & 0.378
           & 0.248
           & 21.261
           & 0.991
           & 0.089
           \\
  &Large   & 0.333
           & 0.465
           & 0.217
           & 24.306
           & \textbf{  0.999}
           & 0.109
           & 0.335
           & 0.455
           & 0.216
           & 23.770
           & \textbf{  0.998}
           & 0.113
           \\ \midrule
  
  \multirow{2}{*}{\makecell{DG-Font}}  
  &Small   & 0.295
           & 0.546
           & 0.197
           & 23.383
           & 0.955
           & 0.067
           & 0.301
           & 0.530
           & 0.198
           & 22.689
           & 0.973
           & 0.040
           \\
  &Large   & 0.301
           & 0.540
           & 0.199
           & 9.914
           & 0.921
           & 0.061
           & 0.306
           & 0.532
           & 0.196
           & 9.819
           & 0.954
           & 0.039
           \\ \midrule
  
  \multirow{2}{*}{\makecell{FsFont}}  
  &Small   & 0.306
           & 0.493
           & 0.235
           & 19.129
           & 0.946
           & 0.010
           & 0.305
           & 0.501
           & 0.222
           & 15.594
           & 0.950
           & 0.024
           \\
  &Large   & 0.318
           & 0.462
           & 0.261
           & 33.084
           & 0.276
           & 0.067
           & 0.327
           & 0.426
           & 0.284
           & 38.676
           & 0.193
           & 0.035
           \\ \midrule
  
  \multirow{2}{*}{\makecell{CG-GAN}}  
  &Small   & 0.314
           & 0.521
           & 0.201
           & 18.929
           & 0.968
           & 0.170
           & 0.317
           & 0.513
           & 0.202
           & 19.210
           & 0.968
           & 0.166
           \\
    &Large  & - & - & - & - & - & - & - & - & - & - & - & - \\ \midrule
  
  \multirow{2}{*}{\makecell{CF-Font}}  
  &Small   & 0.275
           & 0.582
           & 0.183
           & 21.834
           & 0.965
           & 0.111
           & 0.275
           & 0.580
           & 0.181
           & 21.551
           & 0.960
           & 0.121
           \\
  &Large   & 0.269
           & 0.577
           & 0.193
           & 15.829
           & 0.910
           & 0.060
           & 0.271
           & 0.570
           & 0.192
           & 15.270
           & 0.914
           & 0.061
           \\ \midrule
  
  \multirow{2}{*}{\makecell{VQ-Font}}  
  &Small   & 0.274
           & 0.574
           & 0.215
           & 24.552
           & 0.900
           & 0.017
           & 0.278
           & 0.562
           & 0.215
           & 24.563
           & 0.892
           & 0.019
           \\
  &Large   & 0.270
           & 0.584
           & 0.213
           & 19.130
           & 0.880
           & 0.028
           & 0.274
           & 0.572
           & 0.213
           & 19.313
           & 0.872
           & 0.028
           \\ \midrule
  
  \multirow{2}{*}{\makecell{Diff-Font}}  
  &Small   & 0.322
           & 0.515
           & 0.212
           & 7.665
           & 0.868
           & 0.078
           & - & - & - & - & - & - \\
  &Large   & 0.296
           & 0.566
           & 0.178
           & 3.799
           & 0.990
           & 0.194
           & - & - & - & - & - & -  \\ \midrule
  

  \multirow{2}{*}{\makecell{FontDiffuser}}  
  &Small   & 0.326
           & 0.479
           & 0.213
           & 12.513
           & 0.991
           & 0.136
           & 0.329
           & 0.469
           & 0.214
           & 12.929
           & 0.990
           & 0.136
           \\
  &Large   & 0.309
           & 0.522
           & 0.189
           & 14.264
           & 0.992
           & 0.243
           & 0.313
           & 0.511
           & 0.190
           & 14.625
           & 0.991
           & 0.246
           \\ 
  
  \bottomrule

  \multirow{2}{*}{\makecell{Ours\\(n\_ref=1)}}

  &Small   & 0.279
           & 0.599
           & 0.161
           & 3.438
           & 0.995
           & 0.314
           & 0.286
           & 0.583
           & 0.164
           & 3.541
           & 0.994
           & 0.317
           \\
  &Large   & 0.274
           & 0.610
           & 0.153
           & 2.437
           & 0.990
           & 0.547
           & 0.281
           & 0.594
           & 0.157
           & 2.604
           & 0.990
           & 0.545
           \\ \midrule
           
  \multirow{2}{*}{\makecell{Ours\\(n\_ref=10)}}

  &Small   & 0.273
           & 0.612
           & 0.153
           & 3.034
           & 0.993
           & 0.375
           & 0.281
           & 0.595
           & 0.157
           & 3.115
           & 0.993
           & 0.377
           \\
  &Large   & 0.266
           & 0.627
           & 0.142
           & 1.780
           & 0.989
           & 0.642
           & 0.273
           & 0.611
           & 0.147
           & 1.964
           & 0.989
           & 0.634
           \\ \midrule
  
  \multirow{2}{*}{\makecell{Ours\\(n\_ref=100)}}
  &Small   & 0.258
           & 0.643
           & 0.134
           & 2.340
           & 0.997
           & 0.452
           & 0.267
           & 0.623
           & 0.139
           & 2.472
           & 0.996
           & 0.448
           \\
  &Large   & 0.250
           & 0.658
           & 0.125
           & \underline{1.392}
           & 0.993
           & \underline{0.676}
           & 0.259
           & 0.639
           & 0.130
           & \underline{1.581} 
           & 0.991
           & \underline{0.667}
           \\ \midrule
  
  \multirow{2}{*}{\makecell{Ours\\(n\_ref=775)}}
  &Small   & \underline{0.243} 
           & \underline{0.672} 
           & \underline{0.118} 
           & 1.836
           & \underline{0.998}
           & 0.530
           & \underline{0.252} 
           & \underline{0.654} 
           & \underline{0.122} 
           & 1.967
           & \underline{0.997}
           & 0.528
           \\
  &Large   & \textbf{0.234} 
           & \textbf{0.690} 
           & \textbf{0.109} 
           & \textbf{1.209} 
           & 0.997
           & \textbf{0.705} 
           & \textbf{0.243} 
           & \textbf{0.672} 
           & \textbf{0.113} 
           & \textbf{1.400} 
           & 0.996
           & \textbf{0.694} 
           \\ 

           
  \bottomrule
\end{tabular}
\end{center}
\bigskip\centering

\end{minipage}
\end{table*}%

\section{Experiments}

\subsection{Experimental Setup}

\subsubsection{Datasets}
\label{sec:dataset}

To test our model and the compared methods' abilities to model and scale to larger datasets, we construct a small-scale dataset and a large-scale dataset, the former consisting of 6,763 Chinese characters from the full GB2312 standard rendered in 438 fonts, and the latter consisting of 6,763 Chinese characters rendered in 3,538 fonts. The small dataset is a subset of the large one. We randomly extract the same 38 test fonts out of both datasets and leave the rest 400 fonts and 3,500 fonts for training. We leave OptSet out of both training and testing character sets. During training, we randomly select 5,500 characters as the training character set. For testing, we only test on unseen fonts to showcase the models' abilities to generalize to unseen fonts. Specifically, we use randomly selected 640 characters out of the training character set for testing generation quality on seen characters, and use the rest 488 characters for testing generation quality on unseen characters. The same applies for the other compared models as well unless otherwise noted. For quantitative evaluation of our super-resolution models, we report results on a randomly selected subset of the 640 seen characters of size 160 for faster evaluation.  

\subsubsection{Evaluation Metrics}

We use RMSE, SSIM, and LPIPS to evaluate pairwise generation quality, i.e., how similar each generated glyph image is with the corresponding ground truth, and FID to evaluate the distribution-wise similarity between real and generated glyph images. Additionally, following~\cite{park2021few}, we train a character classifier and a font classifier to classify 6,763 classes of characters and 3,538 classes of fonts up to accuracy 99.82\% and 93.52\%, respectively, and test their accuracy on generated images as content accuracy (Acc(C)) and style accuracy (Acc(S)). It should be noted that these two metrics do not precisely represent content and style accuracy. All generated glyph images are resized to a resolution of 64 $\times$ 64 for fair quantitative comparison.

\subsubsection{Implementation Details}
\label{sec:implementation}

For low-resolution model, we train our model for 50 epochs on the small dataset with a batch size of 64, and 20 epochs on the large dataset with a batch size of 128. 
\textcolor{mycolor}{To further facilitate generation in scenarios where only very few references are available, in the last 10\% of the training iterations, we replace the style references with $k$ identical glyph images randomly selected from the reference set with a probability of $\hat{p}$, intending to encourage learning of one-shot generation. In this way, the probability of using component-level character-reference mappings becomes $1-p-\hat{p}$, where $p$ is the probability of using stroke-level character-reference mappings. We set both $p$ and $\hat{p}$ to $0.1$.}
By default, we use a DDPM~\cite{ho2020denoising} sampler to sample for 10 steps, a classifier-free guidance scale of $s_c=2.0$ and $s_s=2.0$, and we use the trailing strategy specified in~\cite{lin2024common} for time step selection. To train the methods being compared, we utilize their default settings on the small dataset. For the larger dataset, we double the number of iterations used for the small one. If further training does not lead to better metric scores, we report metric scores using the default number of training iterations. For one-step distillation, we use our model trained on the large dataset as the teacher model, set classifier-free guidance scales to $s_c=2.0$ and $s_s=2.0$, train for 2 epochs, and adopt a uniform weighting function. For the super-resolution models, we train them using weights initialized from the low-resolution model and use the same sampling settings as those in the low-resolution model. After getting the high-resolution (i.e., 1024 $\times$ 1024) glyph images, we apply the Image Trace tool in Adobe Illustrator to convert them into high-quality vector glyphs (see Fig.~\ref{fig:teaser} and Fig.~\ref{fig:vector_glyphs}).

In addition, we utilize three types of data augmentations specifically designed for glyph images: zooming in or out, widening or narrowing either horizontally or vertically, and applying italic styles. The above-mentioned operations have proven to greatly enhance our method's effectiveness.


\subsection{Comparison with State-of-the-art Methods}

We first compare our Stage A low-resolution model with 8 state-of-the-art methods, including 6 GAN-based models: MX-Font~\cite{park2021multiple}, DG-Font~\cite{xie2021dg}, FsFont~\cite{tang2022few}, CG-GAN~\cite{kong2022look}, CF-Font~\cite{wang2023cf}, VQ-Font~\cite{pan2023few}, and 2 diffusion-based models: Diff-Font~\cite{he2022diff} and FontDiffuser~\cite{yang2023fontdiffuser}. We use the default settings in their papers and source codes, including training hyper-parameters, number of references, image resolution, sampling steps, etc. We train each model on both the small and large datasets, then report test results for both the seen and unseen character sets. However, three exceptions exist: 1) CG-GAN~\cite{kong2022look} fails completely when trained on the large dataset, demonstrating the inferiority of GANs in terms of scaling to large datasets; 2) Diff-Font~\cite{he2022diff} is not generalizable to unseen characters, so we test it on seen characters only; aside from that, we expand their training character set with OptSet to enable applications in Section~\ref{sec:color_texture}. 
\textcolor{mycolor}{Since our total number of training iterations (millions) are larger compared to the default setting of the compared methods (hundreds of thousands), we also provide results of the compared methods trained for the same number of iterations as ours in the supplemental materials to avoid possible unfairness brought by different training settings.}

\begin{figure*}[!t]
	\centering 
	\begin{minipage}{.1\textwidth} 
		\centering 
		{MX-Font\vspace{7pt}
		DG-Font\vspace{7pt}
		FsFont\vspace{7pt}
		CF-Font\vspace{7pt}
		VQ-Font\vspace{7pt}
		Diff-Font\vspace{7pt}
		FontDiffuser\vspace{7pt}
		Ours(n\_ref=1)\vspace{7pt}
		Ours(n\_ref=10)\vspace{7pt}
		Ours(n\_ref=100)\vspace{7pt}
		Ours(n\_ref=775)\vspace{6pt}
		Target}
	\end{minipage}
\hspace{0.02\textwidth}
	\begin{minipage}{.85\textwidth} 
		\centering 
		\includegraphics[width=1.\textwidth]{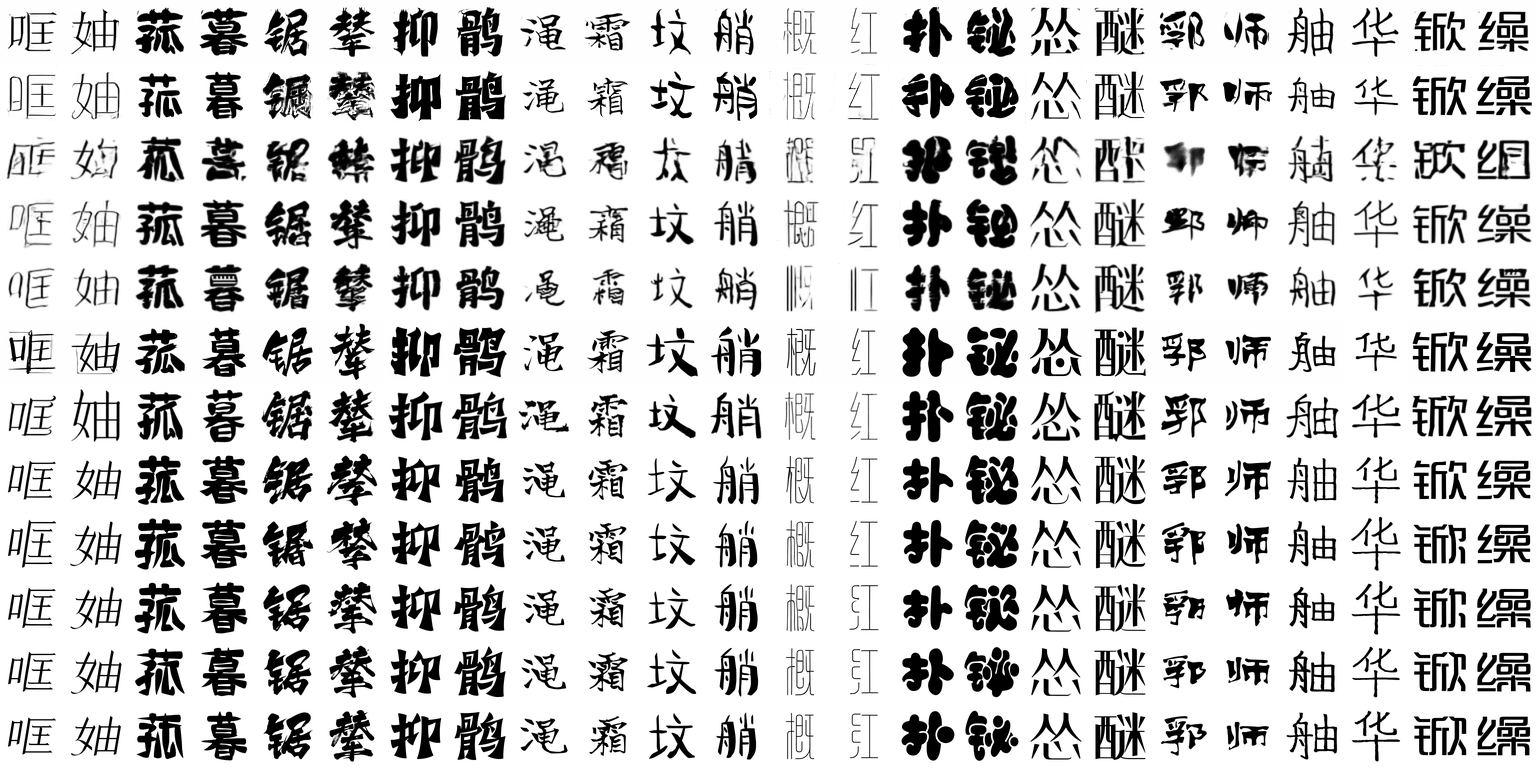}
	\end{minipage} 
	\vspace{-10pt}
	\caption{Comparison of generated results from models trained on the large dataset on unseen fonts seen characters. n\_ref denotes the number of style references.}
	\vspace{-10pt}
 \label{fig:ufsc}
\end{figure*}

Quantitative results are shown in Table~\ref{tab:metrics}. For our model using the reference size (denoted n\_ref) of $n=1,10,100$, we use a size $n$ subset of OptSet as the reference set and reconstruct new mappings accordingly. Except for Acc(C), where the top performing methods all perform similarly, we surpass previous methods across all metrics by a large margin. Notably, using the full OptSet as the reference set, our model achieves a FID of 1.305 on seen characters and a FID of 1.440 on unseen characters. Even using as few as 10 references, our model achieves state-of-the-art results. While Diff-Font~\cite{he2022diff} achieves a relatively low FID of 3.799, it underperforms in other metrics. From their synthesis results, it is evident that despite producing sharp and plausible glyph images, Diff-Font fails to learn a consistent and accurate style (see Fig.~\ref{fig:ufsc}).

\begin{figure*}[!t]
	\centering 
	\begin{minipage}{.1\textwidth} 
		\centering 
		{MX-Font\vspace{7pt}
		DG-Font\vspace{7pt}
		FsFont\vspace{7pt}
		CF-Font\vspace{7pt}
		VQ-Font\vspace{7pt}
		FontDiffuser\vspace{7pt}
		Ours(n\_ref=1)\vspace{7pt}
		Ours(n\_ref=10)\vspace{7pt}
		Ours(n\_ref=100)\vspace{7pt}
		Ours(n\_ref=775)\vspace{6pt}
		Target}
	\end{minipage}
\hspace{0.02\textwidth}
	\begin{minipage}{.85\textwidth} 
		\centering 
		\includegraphics[width=1.\textwidth]{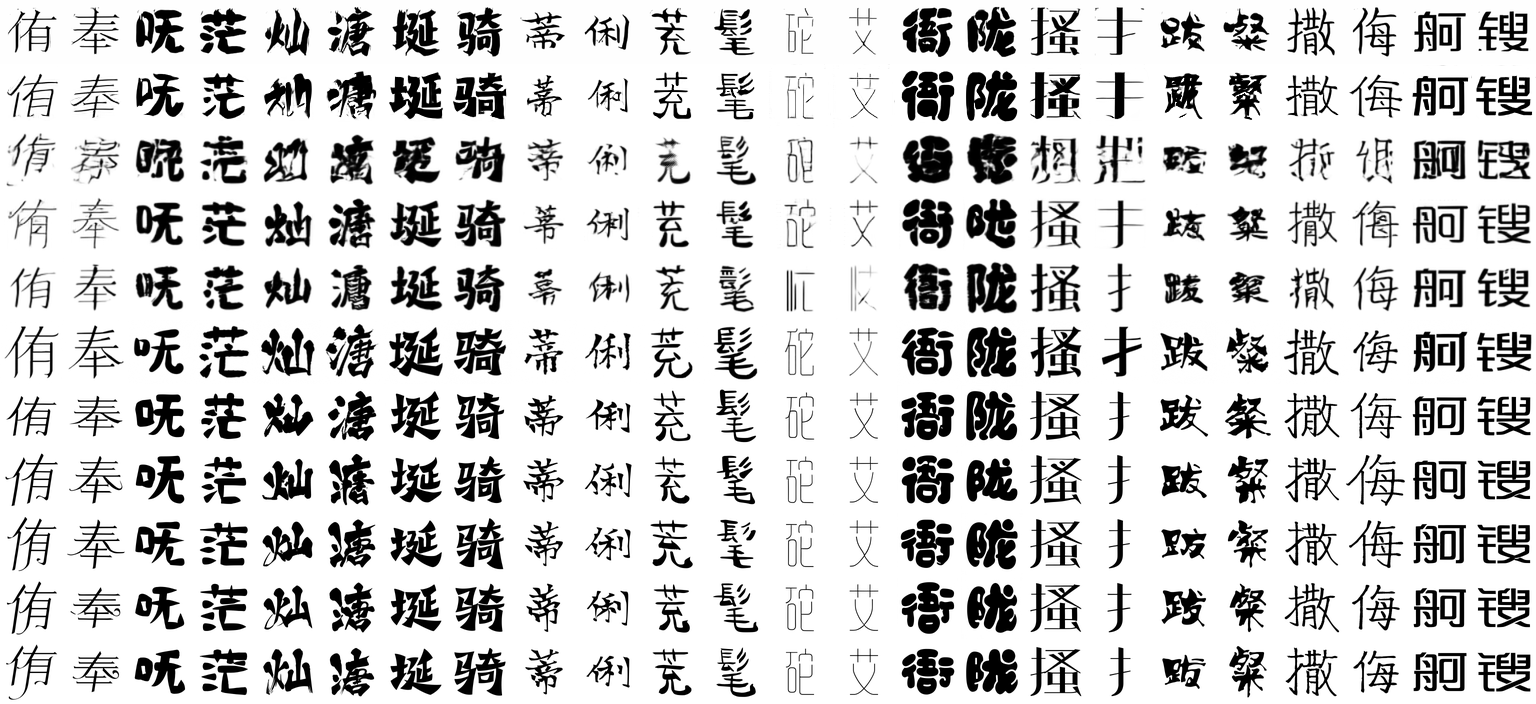}
	\end{minipage} 
	\vspace{-10pt}
	\caption{Comparison of generated results from models trained on the large dataset on unseen fonts unseen characters. n\_ref denotes the number of style references.}
	\vspace{-10pt}
 \label{fig:ufuc}
\end{figure*}

\begin{figure*}[!t]
	\centering 
	\begin{minipage}{.1\textwidth} 
		\centering 
		{
		Ours(n\_ref=10)\vspace{7pt}
		Ours(n\_ref=100)\vspace{7pt}
		Ours(n\_ref=775)\vspace{6pt}
		Target}
	\end{minipage}
\hspace{0.02\textwidth}
	\begin{minipage}{.85\textwidth} 
		\centering 
		\includegraphics[width=1.\textwidth]{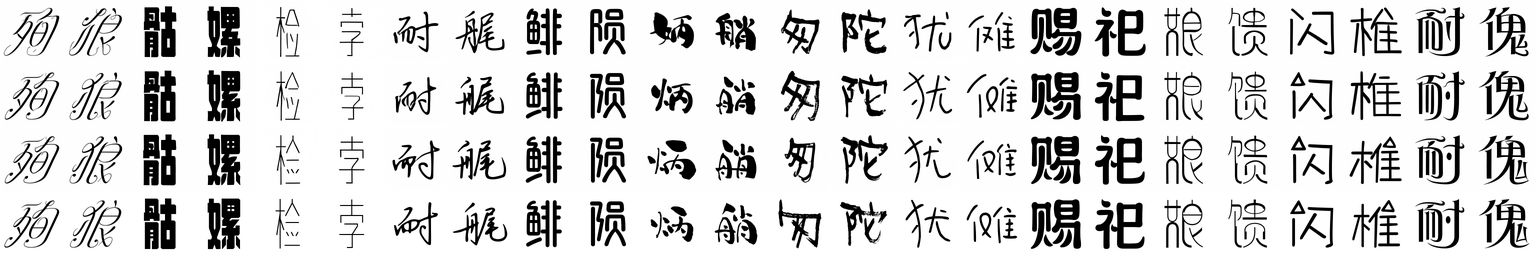}
	\end{minipage} 
	\vspace{-10pt}
	\caption{Extra generated results from our model trained on the large dataset. n\_ref denotes the number of style references.}
	\vspace{-10pt}
 \label{fig:extra}
\end{figure*}

\begin{figure*}[!t]
	\centering 
	\begin{minipage}{.14\textwidth} 
		\centering 
		{
		Original(10 steps)\vspace{7pt}
		Original(1 step)\vspace{7pt}
		Distilled(1 step)\vspace{6pt}
		Target}
	\end{minipage}
\hspace{0.01\textwidth}
	\begin{minipage}{.84\textwidth} 
		\centering 
		\includegraphics[width=1.\textwidth]{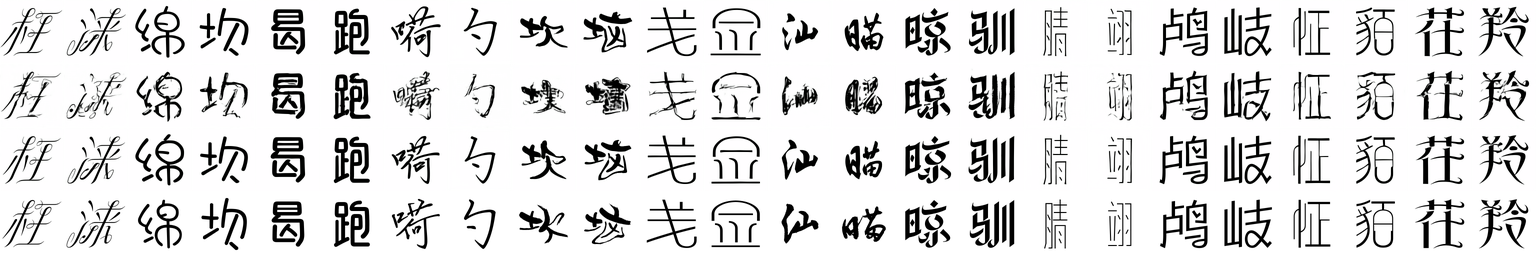}
	\end{minipage} 
	\vspace{-10pt}
	\caption{Comparison of generated results from our one-step distilled model and its multi-step teacher model. The full OptSet is used as the reference set.}
	\vspace{-10pt}
 \label{fig:distill}
\end{figure*}

\begin{table*}[!t]
\caption{Quantitative evaluation of our one-step distilled model and the teacher model on unseen fonts. The full OptSet is used as the reference set.}
\label{tab:distill}
\begin{center}
\npdecimalsign{.}
\nprounddigits{3}
\begin{tabular}{c|c|cccccc|cccccc}
  \toprule

\multirow{2}{*}{Method} & \multirow{2}{*}{\makecell{\#Steps}} & \multicolumn{6}{c|}{Unseen Fonts Seen Characters} & \multicolumn{6}{c}{Unseen Fonts Unseen Characters} \\ &\  &\  RMSE$\downarrow$ & SSIM$\uparrow$ & LPIPS$\downarrow$  & FID$\downarrow$ &  Acc(C)$\uparrow$ & Acc(S)$\uparrow$ & RMSE$\downarrow$ & SSIM$\uparrow$ & LPIPS$\downarrow$  & FID$\downarrow$ & Acc(C)$\uparrow$ & Acc(S)$\uparrow$ \\ \midrule

 \makecell{Teacher}

  &10      & 0.234 
           & 0.690 
           & 0.109 
           & \textbf{1.209} 
           & \textbf{0.997}
           & \textbf{0.705} 
           & 0.243 
           & 0.672 
           & 0.113 
           & \textbf{1.400} 
           & \textbf{0.996}
           & \textbf{0.694} 
           \\   \midrule 

 \makecell{Teacher}

  &1   & 0.230
           & 0.678
           & 0.186
           & 50.882
           & 0.719
           & 0.054
           & 0.238
           & 0.660
           & 0.190
           & 50.497
           & 0.717
           & 0.052
           \\ \midrule 

  \makecell{Distilled}
  &1   & \textbf{0.229}
           & \textbf{0.699}
           & \textbf{0.105}
           & 1.683
           & 0.996
           & 0.675
           & \textbf{0.237}
           & \textbf{0.684}
           & \textbf{0.109}
           & 1.871
           & 0.996
           & 0.669
           \\
           
  \bottomrule
\end{tabular}
\end{center}
\bigskip\centering


\end{table*}%

Qualitative results from models trained on the large dataset are visualized in Fig.~\ref{fig:ufsc} and Fig.~\ref{fig:ufuc}. Please refer to our supplemental materials for synthesis results obtained from various models trained on the small dataset. It can be seen from the generated results that our model produces samples with fine and sharp details while being consistent with the reference images for a variety of styles. Using different sizes of input inferences, it is able to extract different levels of style information from the reference set, e.g., stroke thickness, spatial layout, etc. from a smaller reference set, and delicate brush details, specially designed components, etc. from a larger reference set. The GAN-based methods often exhibit blurriness and artifacts, while the two diffusion-based models produce visually pleasing glyph images thanks to the generative ability of diffusion models but fail to effectively transfer style from input references.




We show extra generated results from our model trained on the large dataset on more difficult cases in Fig.~\ref{fig:extra}, i.e., unseen fonts with complicated styles, special designs, etc. Our model displays remarkable style-wise diversity and consistency as well as impressive content-wise accuracy. In the rest of the paper, if not pointed out particularly, we conduct experiments and present results using our model trained on the large dataset with the full OptSet as the reference set by default.

\subsection{One-step Generation Results}

In this section, we evaluate the effectiveness of our distillation method by comparing our one-step distilled model with the teacher model both quantitatively and qualitatively. We use the same metrics mentioned above to evaluate font generation quality; for the evaluation of generation speed and efficiency, we set the batch size to 64, the image resolution to 64 $\times$ 64, and record the average running time needed to generate the entire seen character test set on a single NVIDIA A40 for all relevant methods.

\begin{figure}[h]
\centering
\includegraphics[width=0.35\paperwidth]{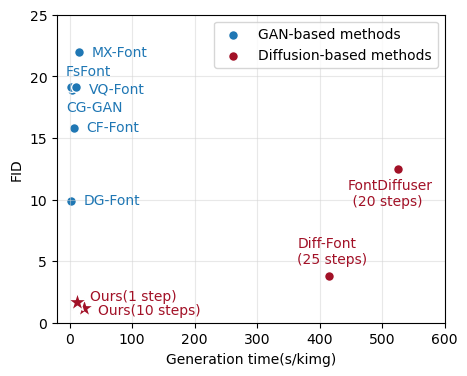}
\caption{Comparison of the performance of different font synthesis methods. Our models maintain an optimal balance between generation quality and efficiency, markedly outperforming other diffusion-based approaches in both two aspects.}
\label{fig:fid_time}
\vspace{-8pt}
\end{figure}

\begin{figure}[!t]
\centering
\includegraphics[width=0.4\paperwidth]{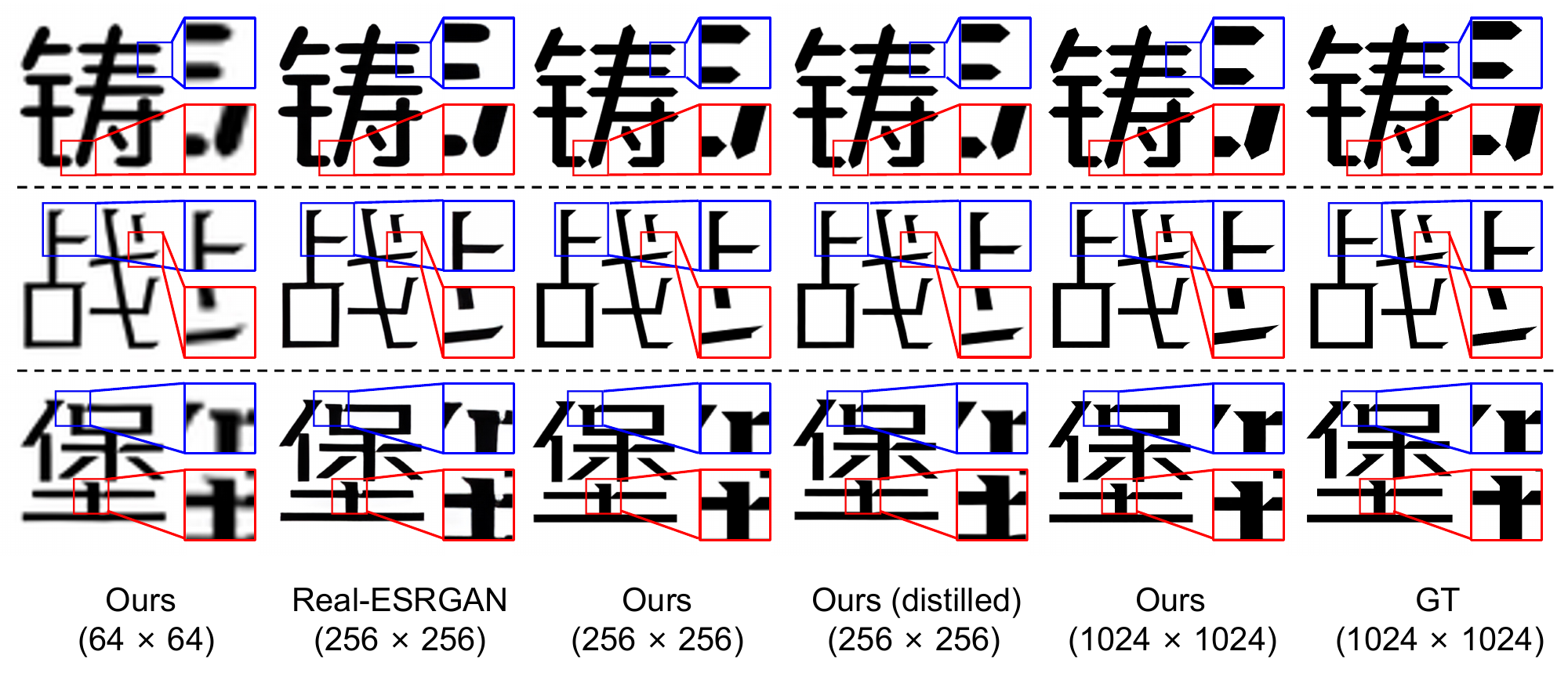}
\caption{Visualization of detailed styles recovered by our super-resolution module.}
\label{fig:highres_detail}
\vspace{-8pt}
\end{figure}

\begin{table}[!t]
\caption{Quantitative evaluation of our SR model compared to an off-the-shelf SR model Real-ESRGAN (denoted as RE). 256 and 1024 denote images in the resolutions of 256 $\times$ 256 and 1024 $\times$ 1024, respectively.}
\label{tab:sr}
\begin{center}
\begin{tabular}{c|cccccc}
  \toprule

Method & RMSE$\downarrow$ & SSIM$\uparrow$ & LPIPS$\downarrow$  & FID$\downarrow$ & Acc(C)$\uparrow$ & Acc(S)$\uparrow$ \\ \midrule

  \makecell{RE\\(256)}   
           & 0.263
           & 0.674
           & 0.123
           & 18.732
           & 0.997
           & 0.637
           \\\midrule
           
  \makecell{Ours\\(256)}   & 0.265
           & 0.676
           & 0.122
           & 14.861
           & 0.996
           & 0.766
              
           \\\midrule
           
  \makecell{RE\\(1024)}   & 0.285
           & 0.657
           & 0.129
           & 15.315 
           & 0.997
           & 0.634
              
           \\\midrule
           
  \makecell{Ours\\(1024)}   & 0.286
           & 0.663
           & 0.127
           & 5.996
           & 0.996
           & 0.742
              
           \\

\bottomrule
\end{tabular}
\end{center}
\bigskip\centering

\end{table}%

Table~\ref{tab:distill} presents the quantitative results of our one-step distilled model and the teacher model. Our distilled model achieves even better RMSE, SSIM and LPIPS results compared to its teacher model, in the meantime achieving decent FID, Acc(C) and Acc(S) scores. Fig.~\ref{fig:distill} presents some generation results of relevant models. We can see that our one-step distilled model faithfully mimics the output of the multi-step teacher model, while the teacher model does not produce valid results in the one-step setting, proving the effectiveness of our design.

Fig.~\ref{fig:fid_time} compares the generation efficiency of our model with other GAN-based and diffusion-based models. It can be observed that our distilled model is significantly faster than the compared diffusion-based models, and comparable with GAN-based methods while achieving much better FID score. As the number of sampling steps drops to as few as 1, the main overhead of the model becomes the decoder part of the autoencoder. Further optimization can be done to improve the efficiency of the autoencoder, which we leave to future work.

\begin{figure*}[!h]
\centering
\includegraphics[width=0.83\paperwidth]{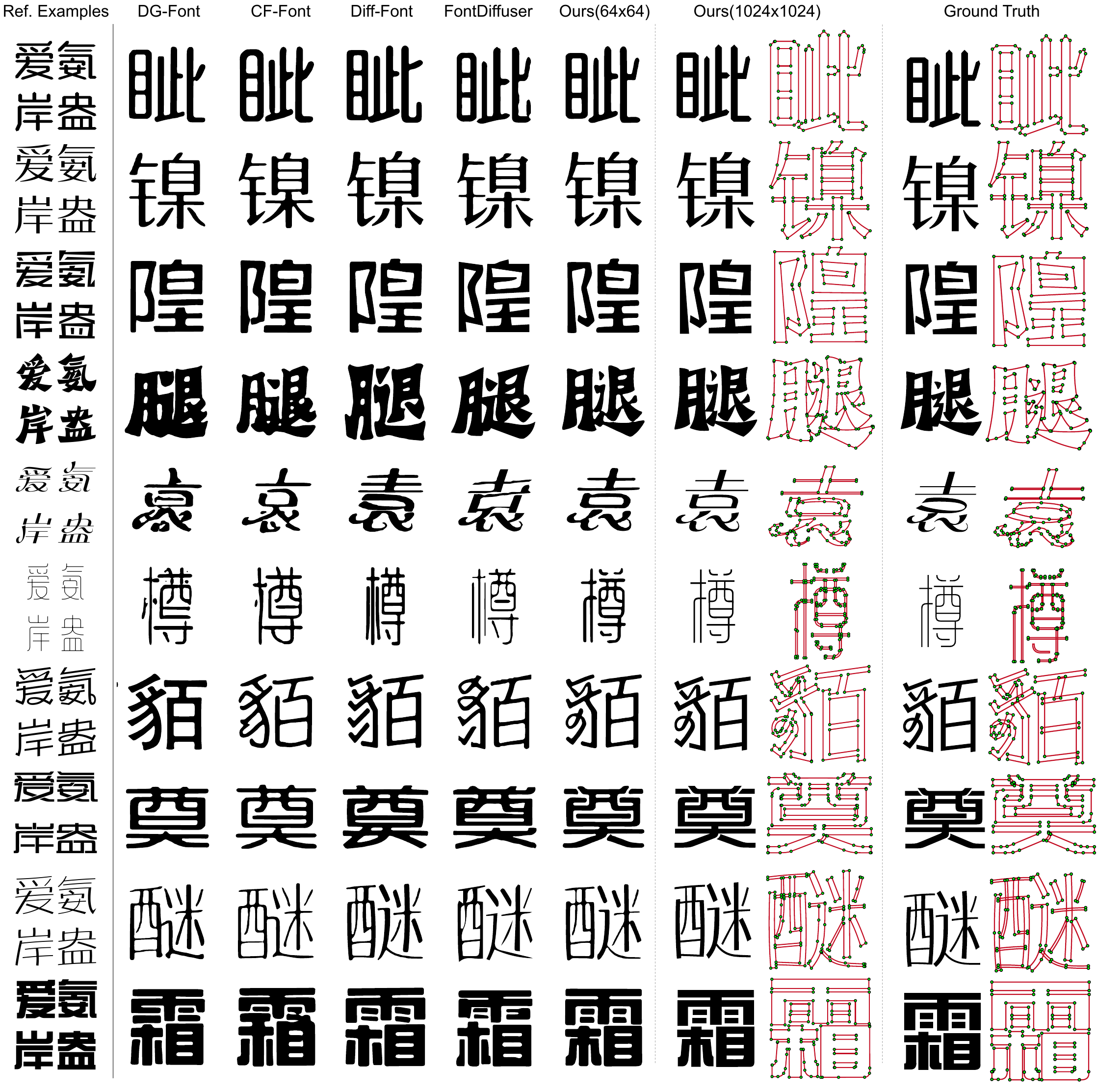}
\caption{Examples of vector glyphs obtained by vectorizing glyph images synthesized by different methods. Four samples of input style references in each font are shown in the first column. The resolution of the original raster images generated by our method is indicated in the figure. Please zoom in for better clarity.}
\label{fig:vector_glyphs}
\end{figure*}

\subsection{High-resolution Generation and Vectorization Results}

In this section, we present high-resolution synthesis results obtained by our super-resolution models. We first compare our model with an off-the-shelf super-resolution model Real-ESRGAN~\cite{wang2021real} in Fig.~\ref{fig:highres_detail} and Table~\ref{tab:sr} to showcase our model’s capability to recover lost style details in low-resolution results. It can be seen clearly from the zoom-in images in Fig.~\ref{fig:highres_detail} that our model is able to extract and recover style details from high-resolution style references. This is also reflected in the increased value of Acc(S).

\textcolor{mycolor}{We have performed the distillation procedure on the super-resolution models and the super-resolution results of the distilled models are shown alongside their multi-step teacher models in Fig.~\ref{fig:highres_detail}.}

\begin{figure*}[!t]
\centering
\includegraphics[width=0.83\paperwidth]{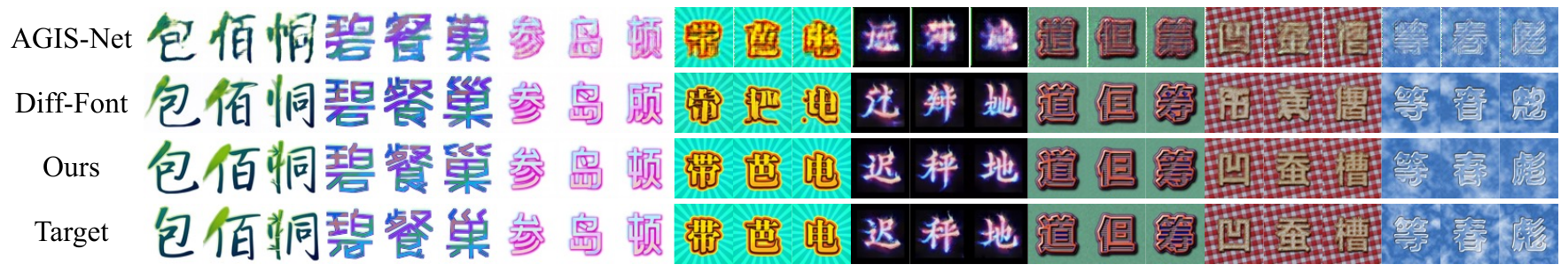}
\caption{Comparison of artistic glyph image synthesis results.}
\label{fig:ct_comp}
\end{figure*}

Generating high-quality glyph images at such high resolution (i.e., 1024 $\times$ 1024) allows us to convert them into vector fonts through a vectorization process with negligible quality loss to narrow the gap between the image modality and the vector modality. We use the Image Trace tool in Adobe Illustrator as an off-the-shelf vectorizer to transform low-resolution glyph images synthesized by our model and several compared models, as well as high-resolution 
glyph images produced by our model into vector glyphs. Vectorization results of glyph images synthesized by different methods are shown in Fig.~\ref{fig:vector_glyphs}. Directly applying vectorization to glyph images generated by current state-of-the-art methods does not yield satisfactory outcomes, as their raster images lack both style accuracy and high-resolution details. On the other hand, our component-aware low-resolution generation model coupled with a style-recovering super-resolution module can successfully produce samples that are of sufficient quality for a vectorizer to produce satisfactory vector glyphs with precise control point positions. To the best of our knowledge, until now, no method directly targeting the automatic generation of high-quality vector fonts has been reported capable of handling glyphs at such level of structural complexity and style diversity. In this paper, we choose an alternative strategy to generate vector fonts by vectorizing high-quality, high-resolution synthesized glyph images, achieving impressive results.




\subsection{Application to Artistic Glyph Image Synthesis}
\label{sec:color_texture}

\begin{figure}[!h]
\centering
\includegraphics[width=0.39\paperwidth]{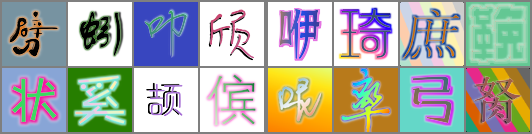}
\caption{Examples of our synthetic colored and textured glyph images.}
\label{fig:ct_aug}
\end{figure}

Artistic glyph image synthesis aims to generate the color and texture of an artistic font in the meantime of generating the shape. It is mostly done with a fine-tuning step, since there are limited training datasets available for the model to learn a representation that can be generalized to unseen patterns, and the possibilities for artistic designs are extensive (e.g., one can replace strokes with some real-world objects, etc.). To equip our model with the ability to deal with colored images, following~\cite{gao2019artistic}, we first pre-train our model with a synthesized colored and textured glyph image dataset. It is constructed by applying common artistic text effects such as outlining, shadowing, and coloring to the foreground and background of glyph images (see Fig.~\ref{fig:ct_aug} for some examples). We randomly apply these operations during training as a form of augmentation instead of generating a large-scale dataset beforehand.

We compare our model with two existing methods: AGIS-Net~\cite{gao2019artistic}, a GAN-based method targeting few-shot artistic glyph image synthesis, and Diff-Font~\cite{he2022diff}. The test artistic fonts come from three datasets: synthetic Chinese character dataset in~\cite{gao2019artistic}, TE141K-C in~\cite{yang2020te141k}, and our synthetic dataset. Since each style in TE141K-C only contains 775 characters from OptSet, we fine-tune each style using randomly selected 100 characters and test on the remaining characters. Regarding implementation details, we train our model on the color-augmented dataset with weights initialized from the model trained on black and white glyph images; during testing, for each given style, we fine-tune our autoencoder for 80 epochs and our LDM for 240 epochs. For the compared methods, we adopt the default settings in AGIS-Net to fine-tune their model for 500 epochs and we fine-tune 320 epochs for Diff-Font to match our total fine-tuning epochs. Note that our model does have a slight advantage over the compared methods since we pre-train on glyph images with augmented colors for both the foreground and background.

Results are shown in Fig.~\ref{fig:ct_comp}. AGIS-Net produces blurry results containing the most visible artifacts. Diff-Font works well on some styles while fails completely on some other styles; also, it sometimes yields incorrect contents likely due to overfitting of the reference images. In contrast, our method performs stably better in terms of both content preservation and style consistency for all test styles. Additionally, our method is the only one demonstrating capability in transferring special effects, e.g., design of leaves and bamboos for the first style in Fig.~\ref{fig:ct_comp}, owing to the component-aware conditioning we construct.

\subsection{Ablation Studies}

In this section, we ablate several design choices in our method, including the reference selection strategy in Section~\ref{sec:ldm_comp}, data augmentation described in Section~\ref{sec:implementation}, and the inclusion of different levels of references in Section~\ref{sec:ldm_comp}.

\begin{figure}[!h]
\centering
\includegraphics[width=0.35\paperwidth]{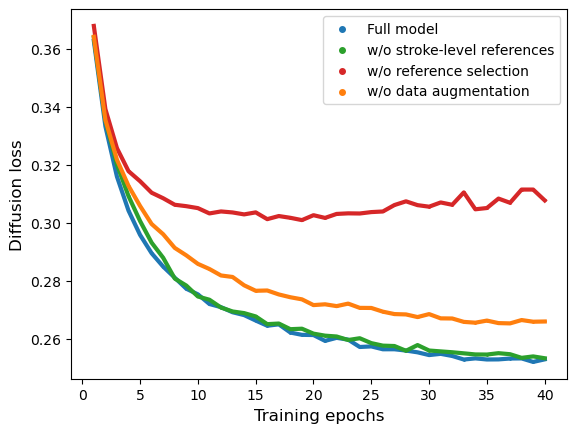}
\caption{The loss curves evaluated on the validation set for different variants of our model.}
\label{fig:loss}
\end{figure}

We first print out the loss curves evaluated on the validation set during training for each variant in Fig.~\ref{fig:loss}. For the variant that excludes reference selection, we use randomly selected $k$ samples as style references. As we can see from the loss curves, the reference selection strategy we use and the data augmentation we incorporate are validated to be beneficial to our model. The more notable improvement is brought by the use of reference selection strategy, indicating its effectiveness in guiding the learning of style transferring. 

\begin{figure}[!h]
\hspace{-0.02\textwidth}
	\centering 
	\begin{minipage}{.1\textwidth} 
		\centering 
		{
		\vspace{1pt}
		w/o stroke\vspace{0.2pt}
		(n\_ref=10)\vspace{2.5pt}\\
		with stroke\vspace{0.2pt}
		(n\_ref=10)\vspace{2.5pt}\\
		w/o stroke\vspace{0.2pt}
		(n\_ref=100)\vspace{2.5pt}\\
		with stroke\vspace{0.2pt}
		(n\_ref=100)\vspace{2.5pt}\\
		Target}
	\end{minipage}
\hspace{-0.01\textwidth}
	\begin{minipage}{.37\textwidth} 
		\centering 
		\includegraphics[width=1\textwidth]{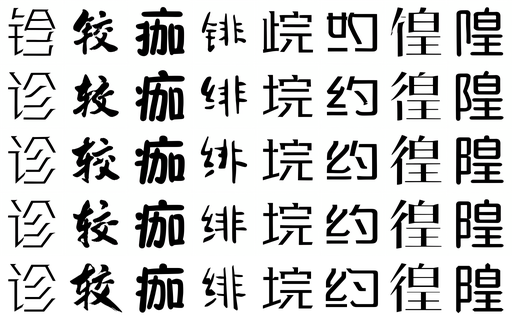}
	\end{minipage} 
	\caption{Our models, without the guidance of stroke-level references, tend to generate incorrect character contents and overlook stylistic details.}
	\vspace{-10pt}
 \label{fig:crstk}
\end{figure}

For evaluation of the effect of using different levels of references, i.e., selecting from both stroke-related and component-related references, we show generated results from the model trained using only component-related references in Fig.~\ref{fig:crstk}. The effect becomes more visible as the reference size decreases, where the model demonstrates a behavior of transferring wrong components to target characters instead of generating results adhering to given contents. We hypothesize that the model only learns a component transfer function during training, and does not learn to adhere to the input character structure, as it is provided with all the necessary components to reconstruct the target glyph. By weakening the input style references, we force the model to adjust to different levels of style information, learn different levels of style representations, and actively stay faithful to the given content. For qualitative results of other variants, we provide them in the supplemental materials.

\subsection{User Studies}
\label{sec:user}

Since quantitative indicators do not strictly align with human perception, we also conduct user studies to further evaluate our method’s effectiveness, which comprises two parts: 1) we perform a Turing test with non-expert participants using our low-resolution synthesis results to find out if our generated glyph images are indistinguishable from ground truths to common users; 2) to evaluate the performance of our method in the real-world font design scenario, we recruit several professional font designers to assess the quality of our generated vector glyphs from an expert point of view. Both user studies are conducted with our model trained on the large dataset with the full OptSet as the reference set. 

\begin{figure*}[!h]
\centering
\includegraphics[width=0.8\paperwidth]{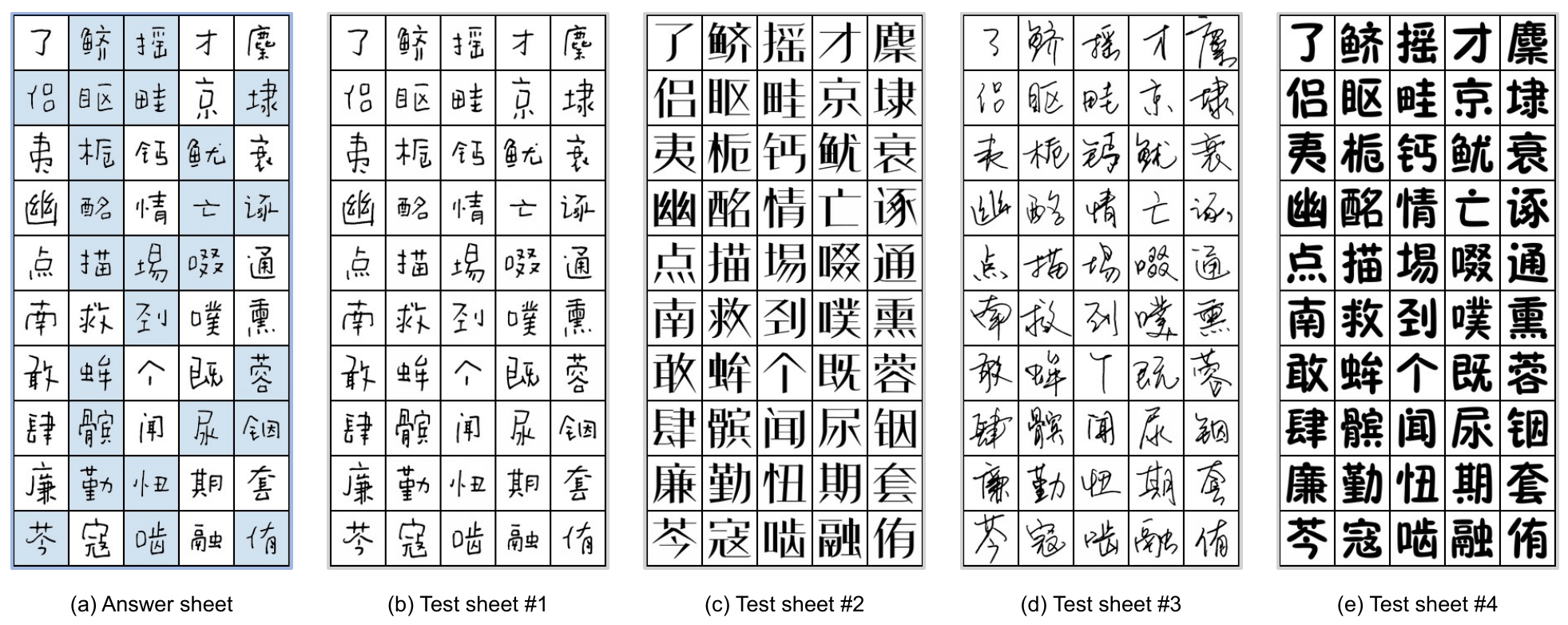}
\caption{Example test sheets (b-e) shown to participants for the first part of our user studies. As shown in the answer sheet (a), glyphs generated by our system are marked in blue and the rest are ground-truth glyph images.}
\label{fig:test_sheet}
\end{figure*}

In the first part of our user studies, the participants are first shown 10 examples from the input reference set, then presented with 25 generated results mixed with 25 ground-truth samples and asked to choose which ones are computer-generated (see Fig.~\ref{fig:test_sheet} for examples). The experiment is conducted on 10 fonts from our test set mentioned in Section~\ref{sec:dataset} and all characters are randomly selected with no duplication. We collect a total of 211 data points from 93 participants. The average accuracy across all participants and all font styles is 51.07\%, which is close to a 50/50 random guess, indicating that our generated results are practically indistinguishable from references to common users. 

\begin{figure*}[!h]
\centering
\includegraphics[width=0.8\paperwidth]{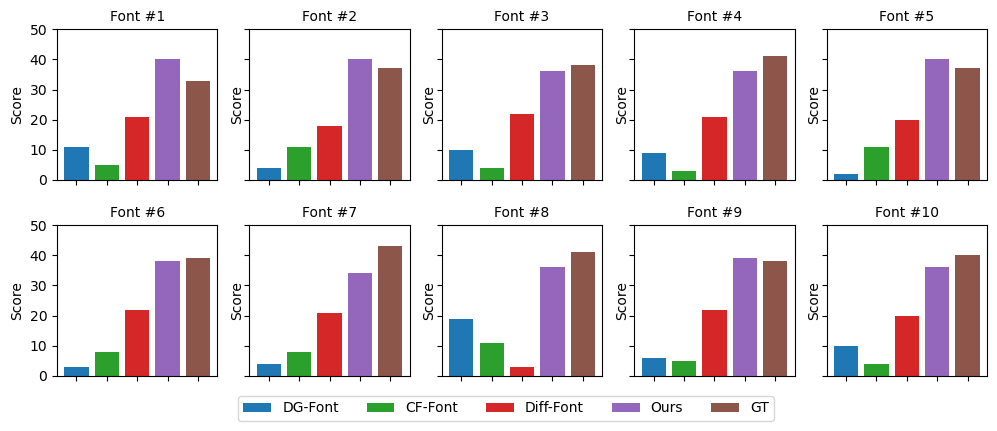}
\caption{Calculated scores for each relevant vector glyph from each font in the second part of our user studies. Higher score denotes higher quality of the corresponding vector glyph (see Section~\ref{sec:user} for details regarding the collection and calculation of scores). We can see that even professional font designers struggle to distinguish ground-truth vector glyphs from those generated by our method, which significantly outperforms other existing approaches.}
\label{fig:user_scores}
\end{figure*}

In the second part of our user studies, professional font designers are shown the 10 groups of vector glyphs in Fig.~\ref{fig:vector_glyphs}, each containing 4 generated vector glyphs and 1 ground-truth vector glyph placed in random order. We use vector glyphs vectorized from our high-resolution synthesis results to represent our model. Then, given several corresponding reference vector glyphs, participants are asked to rank the 5 vector glyphs (including the ground truth) from best to worst quality. In total, 11 font designers have participated in our experiment. We assess each font style individually; each generated vector glyph ranked as first to last is assigned a score of 4 to 0, and we calculate the total score for each generated vector glyph across all participants. Results for each font style are compared in Fig.~\ref{fig:user_scores}. Our model reaches the highest score for 4 out of the 10 fonts and reaches the second highest score for 6 of the fonts, led by the ground-truth vector glyphs only. This suggests that our generated vector fonts are significantly better than those from the compared methods and are even comparable to designer-created ground-truth vector fonts, demonstrating the great potential of our model in automatically fulfilling the task of professional vector font design.

\begin{figure*}[!h]
\centering
\includegraphics[width=0.83\paperwidth]{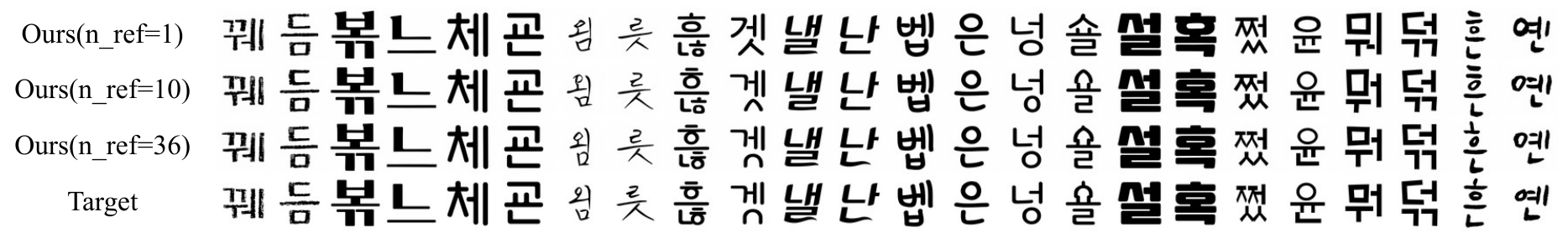}
\caption{Korean glyphs synthesized by models fine-tuned from our Chinese generation model on unseen fonts seen characters. n\_ref denotes the number of style references.}
\label{fig:kor_pretrain_seen}
\end{figure*}

\begin{figure}[!h]
\centering
\includegraphics[width=0.4\paperwidth]{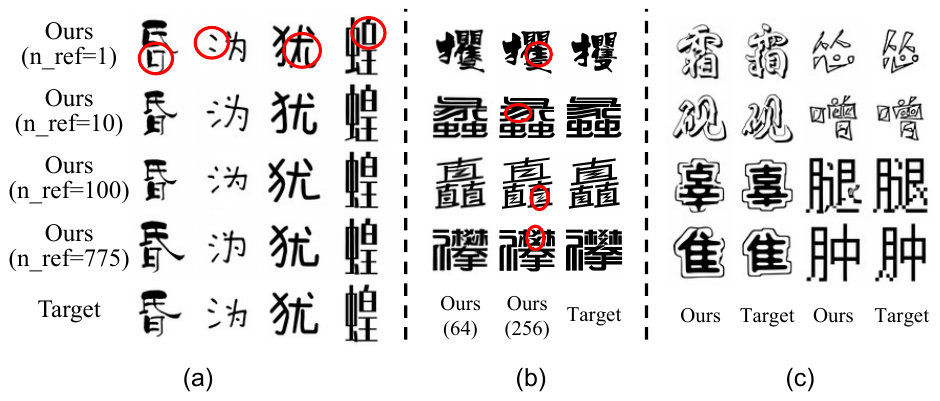}
\caption{Failure cases of our method. Incorrect character structures are marked out with red circles. (a) Content accuracy drops as the number of references decreases. (b) Blurry, duplicated or missing strokes, which cannot be fixed by our SR model. (c) Imperfect style transfer.
}
\label{fig:failure_cases}
\end{figure}

\section{Failure Cases and Limitations}

In this section, we present the failure cases of our model and discuss corresponding limitations.

Firstly, our model does not have explicit control over character structure, impairing its robustness in terms of content accuracy, especially with fewer input references or for complicated characters. As we can see from Table~\ref{tab:metrics} and Fig.~\ref{fig:failure_cases} (a), the content accuracy drops as the number of references decreases. Even when using the full OptSet as the reference set, as depicted in Fig.~\ref{fig:failure_cases} (b), for a few Chinese characters with complicated shapes, our model sometimes generates glyphs with duplicated or missing strokes, or synthesizes blurry results with densely placed strokes, which cannot be fixed by our super-resolution models. Efforts can be put into equipping our super-resolution models with prior knowledge on character structures to recover lost contents that are difficult to generate in low-resolution images. Apart from that, incorporating advanced content preserving techniques \cite{yang2023fontdiffuser} into our model may also be helpful . 

Secondly, although our model demonstrates impressive style learning and transferring capability, there are still corner cases that our model cannot handle. Cases of imperfect style transferring are shown in Fig.~\ref{fig:failure_cases} (c). Our model sometimes confuses the shadow or the outline parts with the actual glyph. Conducting augmentations of adding shadows or outlines to glyphs may be helpful.


\begin{figure}[!h]
\centering
\includegraphics[width=0.4\paperwidth]{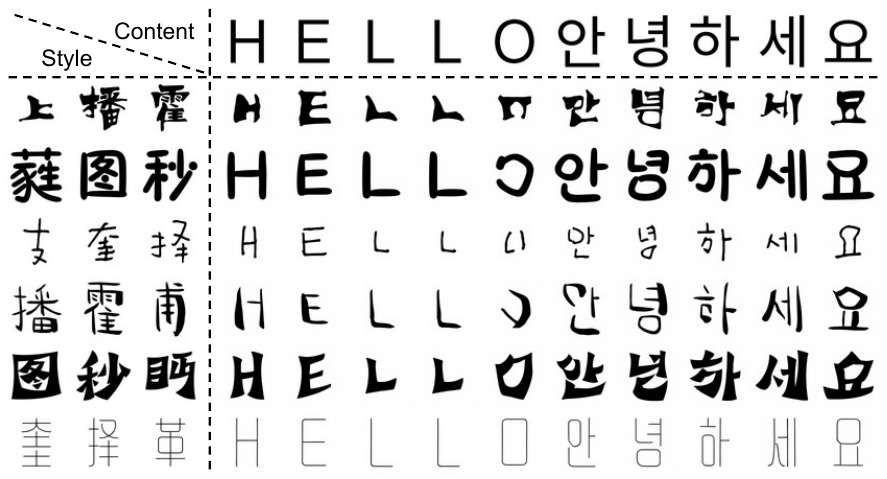}
\caption{Cross-language font generation results of our Chinese font synthesis model without fine-tuning on other languages.}
\label{fig:cross_lan}
\end{figure}

\begin{figure}[!h]
\centering
\includegraphics[width=0.3\paperwidth]{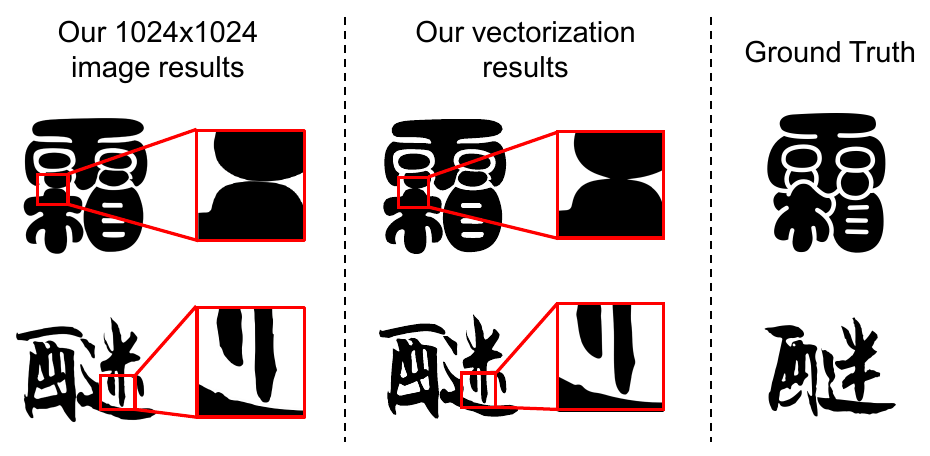}
\caption{Some less satisfactory vectorization results.}
\label{fig:vec_fail}
\end{figure}

Also, our method is based on the structural decomposition of Chinese characters and does not generalize well to other languages, especially for shapes that do not exist in the GB2312 Chinese character set, e.g., the shape of a circle. See Fig.~\ref{fig:cross_lan} for examples of English and Korean characters. In order to achieve cross-language generalization, methods to better grasp a global style to transfer to glyphs from other languages can be added upon our framework. \textcolor{mycolor2}{Nevertheless, our model still possesses valuable knowledge of extracting and transferring styles in glyph images and can serve as a strong initialization for the generation of other languages. We directly apply our Chinese generation framework on Korean characters using our Chinese generation model as weight initialization and present several generated results in Fig.~\ref{fig:kor_pretrain_seen}, where the strong ability of learning multi-level designs are shown to transfer to the generation of Korean characters. Please refer to the supplementary materials for detailed description and discussion of our method’s extension to other languages.}

\textcolor{mycolor2}{Lastly, we discuss the limitation of our vectorization process. The off-the-shelf vectorizer we use is a well-developed commercial tool that we empirically find to perform fairly robustly. However, we did observe some less satisfactory cases. As presented in Fig.~\ref{fig:vec_fail}, the vectorizer sometimes fails to binarize our generated glyph images properly, connecting components that are supposed to be separated (first row), and sometimes produces overly smoothed results that fail to perfectly preserve the original font style (second row). }



\section{Conclusions}

This paper proposed HFH-Font, a few-shot Chinese font synthesis framework capable of generating high-fidelity and high-resolution glyph images that can be vectorized into high-quality vector fonts. Our component-aware encoders and reference selection strategy bring the flexibility of varying reference sizes into our system. Our proposed distillation method and detail-recovering super-resolution module are proven to be effective in improving the efficiency and generation quality of our system. Extensive experiments including user studies have been conducted to show that results produced by our system not only surpass existing state-of-the-art methods, but are even comparable to designer-created ones, raising the possibility of automatic generation of large-scale Chinese vector fonts.


\begin{acks}
This work was supported by National Natural Science
Foundation of China (Grant No.: 62372015), Center For Chinese Font Design and Research, and Key Laboratory of Intelligent Press Media Technology.
\end{acks}

%
%
%
%

\bibliographystyle{ACM-Reference-Format}
\bibliography{myref}


\end{document}